\documentclass[10pt,twocolumn,letterpaper]{article}

\usepackage[pagenumbers]{cvpr} 

\usepackage{graphicx}
\usepackage{amsmath}
\usepackage{amssymb}
\usepackage{booktabs}
\usepackage[accsupp]{axessibility}

\usepackage[pagebackref,breaklinks,colorlinks]{hyperref}

\usepackage[capitalize]{cleveref}
\crefname{section}{Sec.}{Secs.}
\Crefname{section}{Section}{Sections}
\Crefname{table}{Table}{Tables}
\crefname{table}{Tab.}{Tabs.}

\usepackage{graphicx}
\usepackage[table,xcdraw]{xcolor}
\usepackage{multirow}
\usepackage{placeins}

\begin{document}

\title{Principles of Forgetting in Domain-Incremental Semantic Segmentation in Adverse Weather Conditions}
\author{Tobias Kalb\\
Porsche Engineering Group GmbH\\
 Weissach, Germany\\
{\tt\small tobias.kalb@porsche-engineering.de}
\and
J\"urgen Beyerer\\
Fraunhofer IOSB \& Karlsruhe Institute of Technology\\
 Karlsruhe, Germany\\
{\tt\small juergen.beyerer@iosb.fraunhofer.de}
}
\maketitle

\begin{abstract}
Deep neural networks for scene perception in automated vehicles achieve excellent results for the domains they were trained on.
However, in real-world conditions, the domain of operation and its underlying data distribution are subject to change. Adverse weather conditions, in particular, can significantly decrease model performance when such data are not available during training.
Additionally, when a model is incrementally adapted to a new domain, it suffers from catastrophic forgetting, causing a significant drop in performance on previously observed domains.
Despite recent progress in reducing catastrophic forgetting, its causes and effects remain obscure.
Therefore, we study how the representations of semantic segmentation models are affected during domain-incremental learning in adverse weather conditions.
Our experiments and representational analyses indicate that catastrophic forgetting is primarily caused by changes to low-level features in domain-incremental learning and that learning more general features on the source domain using pre-training and image augmentations leads to efficient feature reuse in subsequent tasks, which drastically reduces catastrophic forgetting. 
These findings highlight the importance of methods that facilitate generalized features for effective continual learning algorithms.

\end{abstract}

\section{Introduction}\label{sec:intro}
Semantic segmentation is widely used for environment perception in automated driving, where it aims at recognizing and comprehending images at the pixel level.
One fundamental constraint of the traditional deep learning-based semantic segmentation models is that they are often only trained and evaluated on data collected mostly in clear weather conditions and that they assume that the domain of the training data matches the domain they operate in.
However, in the real world, those autonomous driving systems are faced with constantly changing driving environments and variable input data distributions. Specifically, changing weather conditions can have adverse effects on the performance of segmentation models.

\begin{figure}
\centering
\includegraphics[width=\columnwidth]{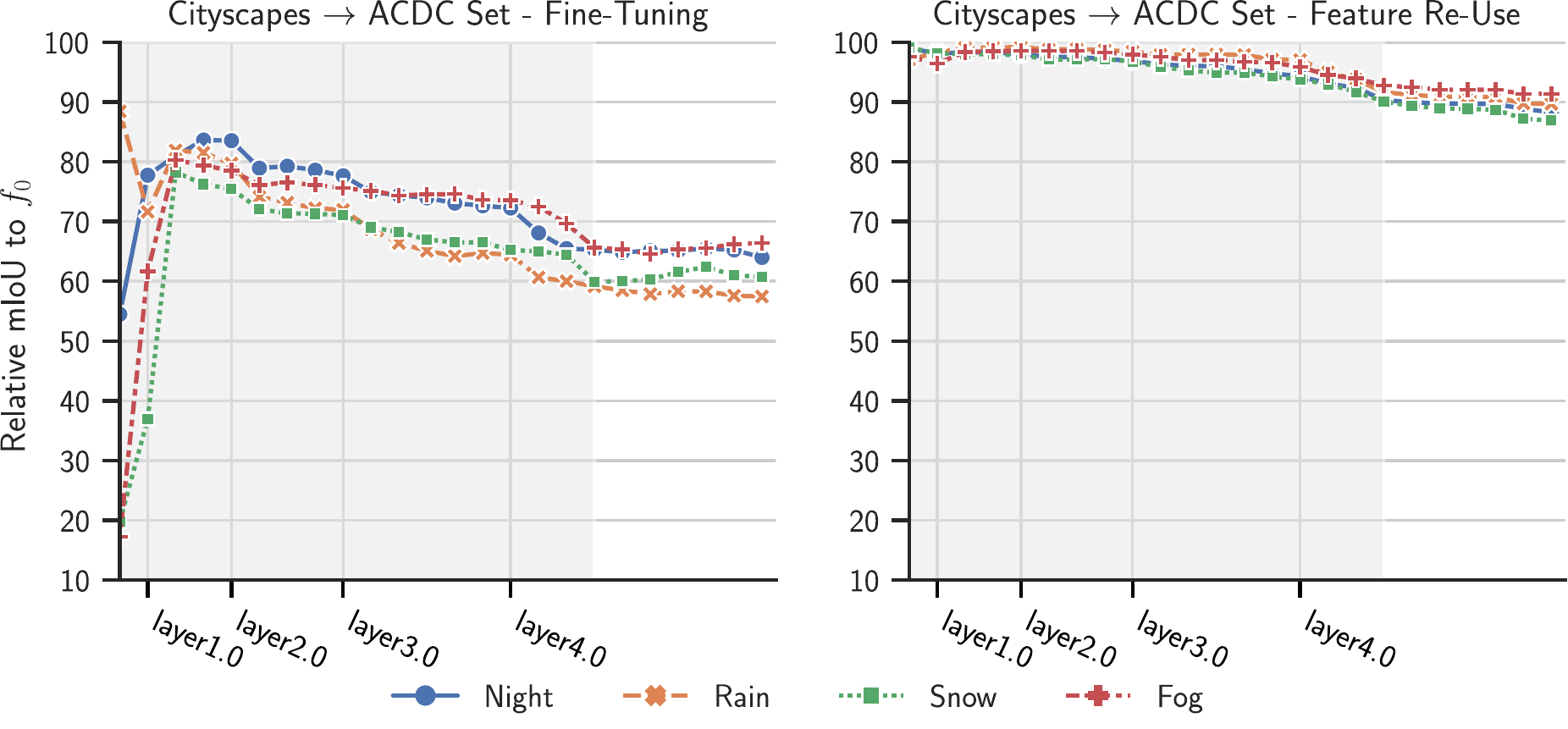}
\caption{Activation drift between models $f_1$ to $f_0$ measured by relative mIoU on the first task of the models stitched together at specific layers (horizontal axis). The layers of the encoder are marked in the gray area, the decoder layers in the white area. Layer-stitching reveals that during domain-incremental learning, changes in low-level features are a major cause of forgetting. With an improved training scheme, combining simple augmentations, exchanging normalization layers and using pre-training, the model is optimized to reuse low-level features during incremental learning, leading to significant reduction of catastrophic forgetting.}
\label{fig:layer_stitch_results}
\end{figure}

Therefore, a semantic segmentation model needs to be adapted to these conditions. A naive solution to this problem would be to incrementally fine-tune the model to new domains with labeled data. 
However, fine-tuning a neural network to a novel domain will, in most cases, lead to a severe performance drop in previously observed domains. 
This phenomenon is usually referred to as \emph{catastrophic forgetting} and is a fundamental challenge when training a neural network on a continuous stream of data. Recently proposed methods mostly mitigate this challenge by replaying data from previous domains, re-estimating statistics or even in an unsupervised manner by transferring training images in the style of the novel domain \cite{Wu_2019_ICCV, cuda_robert,9564566}. 
The focus of our work is to study how the internal representations of semantic segmentation models are affected during domain-incremental learning and how efficient feature reuse can mitigate forgetting without explicit replay of the previous domain.
Our main contributions are:
\begin{enumerate}
    \setlength{\itemsep}{1pt}
    \setlength{\parskip}{0pt}
    \setlength{\parsep}{0pt}
    \setlength{\partopsep}{0pt}
    \item We analyze the activation drift that a model's layers are subjected to when adapting from good to adverse weather conditions by stitching them with the previous task's network. We reveal that the major cause of forgetting is a shift of low-level representations in the first convolution layer that adversely affects the population statistics of the following BatchNorm Layer.
    \item Using different augmentation strategies to match the target domains in color statistics or in the frequency domain, we reveal that learning color-invariant features stabilizes the representations in early layers, as they don't change when the model is adapted to a new domain.
    \item With a combination of pre-training, augmentations and exchanged normalization layers, we achieve an overall reduction of forgetting of $\sim$20\% mIoU compared to fine-tuning without using any form of replay and prove the effectiveness of pre-training and augmentations which are often overlooked in continual learning.
\end{enumerate}

\section{Related Work}
\subsection{Continual Learning}
Continual learning research is primarily concerned with developing methods to overcome catastrophic forgetting. It has been extensively studied in incremental classification tasks, where the approaches can be broadly divided into regularization-based methods\cite{Aljundi2018_MAS,Kirkpatrick2015,aljundi_survey,synaptic_intelligence,Li2018}, replay-based methods \cite{shin2017continual,NEURIPS2019_fa7cdfad,rebuffi-cvpr2017,hayes2020remind} and parameter-isolation methods \cite{mallya2018packnet,yoon2017lifelong,wortsman2020supermasks}. 
For semantic segmentation, significant progress has been made on addressing class-incremental learning, mostly utilizing knowledge distillation-based approaches \cite{Michieli2019,klingner2020class,Cermelli2020,Douillard2020,Michieli2021,Kalb2021,9737321,Cermelli_2022_CVPR}. Research on domain-incremental learning for semantic segmentation is relatively sparse. 
Garg \etal \cite{Garg_2022_WACV} propose a dynamic architecture that learns dedicated parameters to capture domain-specific features for each domain. 
Mirza \etal \cite{mirza2022efficient} circumvent the issues of biased BatchNorm statistics by re-estimating and saving them for every domain, so that during inference domain-specific statistics can be used. Our experiments in this paper show that the effectiveness of these approaches mostly originates in matching the low-level statistics of the initial block of the network. 
Recently, the focus shifted towards Continual Unsupervised Domain Adaptation (CDA).

\subsection{Continual Unsupervised Domain Adaptation}
CDA has the goal of adapting a model that is trained on a supervised source dataset to a sequence of different domains, for which no labels are provided. However, in order to compensate for the missing labels of the target domains, the model has access to the initial source dataset throughout the entire training sequence \cite{cuda_robert}. Methods in this category work mostly by storing information about the style of the specific domains, so that during training the source images can be transferred into the styles of the different target domains. This can be achieved by storing low-frequency components of the domains \cite{9564566} or by capturing the style using generative models \cite{Wu_2019_ICCV, cuda_robert}. Other recent work proposes to use a target-specific memory for every domain to mitigate forgetting \cite{cuda2}.
Although this setting is different from the domain-incremental setting in our work, the insights that we gain on the causes and mitigation strategies, especially the importance of the invariance of low-level features, can be transferred to CDA as well.

\subsection{Analysing Catastrophic Forgetting}
In previous studies, representational analysis techniques such as centered kernel alignment \cite{Kornblith2019} and linear probing were used to analyze the effects of catastrophic forgetting in deep learning for image classification \cite{davari2021probing,ramasesh2021anatomy}. Their results concluded that deeper layers are disproportionately responsible for forgetting in class-incremental learning.
Other work investigates how multi-task- and incremental learning solutions are connected in their loss landscape \cite{mirzadeh2021linear}, or how task sequence \cite{Nguyen2019} or task similarity \cite{ramasesh2021anatomy} affect catastrophic forgetting.
Overall, much of the recent research in analyzing catastrophic forgetting has been primarily focused on classification tasks or class-incremental learning \cite{Kalb2022,ramasesh2021anatomy}.
While much progress has been made on mitigating forgetting in domain-incremental semantic segmentation by utilizing style transfer, replay or matching BN statistics, it remains unclear how these specific changes affect the internal representations of the model.

\section{Preliminaries}
\subsection{Problem Formulation} 
The task of semantic segmentation is to assign a class, out of a set of pre-defined classes $\mathcal{C}$, to each pixel in a given image. 
A training task $T = \{(x_n, y_n)\}^{N}_{n=1}$ consists of a set of $N$ images $x\in \mathcal{X}$ with $\mathcal{X} = \mathbb{R}^{H \times W \times 3}$ and corresponding labels $y\in \mathcal{Y}$ with $\mathcal{Y} = \mathcal{C}^{H \times W}$. 
Given the task $T$ the goal is to learn a mapping $f: \mathcal{X} \mapsto \mathbb{R}^{H\times W\times |\mathcal C|}$ from the image space $\mathcal{X}$ to a score vector $\hat{y}$. The final segmentation mask is then computed as $\bar{y}_i = \operatorname{arg\,max}_{c\in C}\hat{y}_{i, c}$.
In the incremental learning setting, the model $f$ is trained on a sequence of tasks $T_k$ that can introduce new classes or visually distinct instances from the same classes. This study focuses on domain-incremental settings, where the input distribution of the data is changing between tasks while the set of classes is fixed.
Specifically, we focus on domain-increments in different adverse weather conditions.

\subsection{Normalization Layers}
Normalization layers are essential in training Convolutional Neural Networks (CNNs), as they address the internal covariate shift of the network by normalizing the inputs to layers, so that the input distributions to each layer are stable during training \cite{batchnorm}.  
Batch Normalization (BN)~\cite{batchnorm} is the most common CNN normalization layer, which normalizes layer inputs using moments across the mini-batch dimension.
However, to achieve deterministic behavior during inference, the mini-batch variance and mean are replaced by the global population mean and variance, which are obtained during training using an exponential running average. 
This works for i.i.d.\footnote{independent and identically distributed} data, but in non-i.i.d. incremental learning, the BN estimates of the population mean and variance are heavily biased towards the most recent task, resulting in a significant loss of performance on old tasks  \cite{Lomonaco_2020_CVPR_Workshops}. 
Continual Normalization (CN) alleviates this discrepancy by combining Group- \cite{Wu_2018_ECCV} and Batch Normalization \cite{pham2021continual}.

\subsection{Measuring Activation Drift}\label{sec:layer_stitch}
\begin{figure}
\centering
\includegraphics[width=0.8\columnwidth]{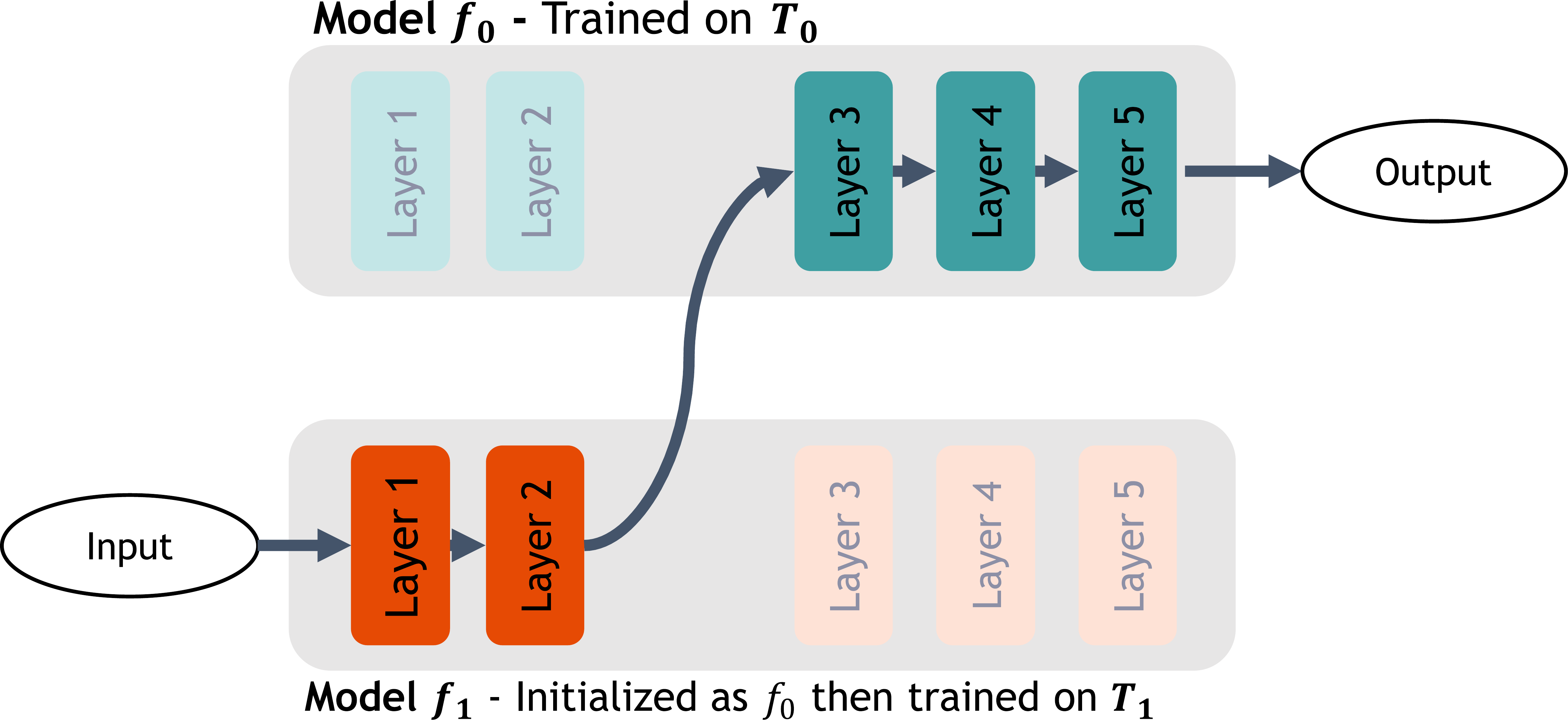}
\caption{Layer stitching: By directly propagating the activations of $f_1$ to $f_0$ we measure the feature reuse between the model $f_0$ and $f_1$ \cite{Kalb2022}.}
\label{fig:layer_stitch_setup}
\end{figure}

Accuracy-based evaluation only allows for restrictive insight into the causes of forgetting of a model, specifically if representation shifts happen in the early layers of the networks. 
As we expect catastrophic forgetting to be the result of a change in weights and activations of the model that are no longer tuned to the most recent task, we aim to measure activation drift in incremental learning. Specifically, we want to measure the activation shift between a model $f_0$ and $f_1$. Where $f_0$ denotes the model trained on $T_0$ and $f_1$ is the model initialized with the parameters of $f_0$ and incrementally trained on $T_1$. 
For this, we utilize the layer matching framework introduced by Csisz\'{a}rik \etal \cite{Csiszarik2021}, without an additional stitching layer~\cite{Kalb2022}. \cref{fig:layer_stitch_setup} shows the layer stitching setup. 
In this setup, we measure the impact of activation drift until layer $n$ by propagating the activations of layer $n$ directly  to the layer $n+1$ in $f_0$. The resulting stitching model is then evaluated on the test set of the initial task $T_0$ and compared to the initial accuracy of the model $f_0$. This gives us a proxy measure of how useful the features at a specific layer of the adapted model $f_1$ are for the initial model $f_0$.

\section{Experiments}
\paragraph{Datasets:} We conduct our experiments on adapting to adverse weather conditions using an domain-incremental setup that involves adapting from the Cityscapes (CS)~\cite{cityscapes} dataset to ACDC~\cite{acdc}, which is commonly used as a benchmark for unsupervised domain adaptation.
The CS dataset is an automotive semantic segmentation dataset, collected during daytime and dry weather conditions in different German, Swiss and French cities. It contains 2975 training and 500 validation images. 
The ACDC dataset is collected during different adverse weather conditions and divided into four different subsets: \textit{Night}, \textit{Rain}, \textit{Fog} and \textit{Snow}. 
ACDC and CS share the same 19 classes, so that the changes between the tasks is only based on the domain differences. 
To study how features are reused and or adapted in each adverse weather condition, we investigate four different scenarios all starting with the same CS model: \textit{CS} $\rightarrow$ \textit{Night}, \textit{CS} $\rightarrow$ \textit{Rain}, \textit{CS} $\rightarrow$ \textit{Snow} and \textit{CS} $\rightarrow$ \textit{Fog}.

\paragraph{Models:} We use the widely adopted DeepLabV3+ \cite{Chen_2018_ECCV} in our experiments with a ResNet50 backbone, as DeepLabV3+ is a commonly used architecture in domain adaptation. Furthermore, we confirm our findings with different architecture in Appendix F, because the architectural choices can have a significant impact in continual learning~\cite{archmatters,kalb2023effects}. 
Finally, in \cref{segformer} we show that the recently introduced SegFormer-B2 \cite{xie2021segformer} is more robust towards low-level feature change than its CNN counterparts. In the majority of the experiments, we use randomly initialized models for training on the first task, as pre-training is known to increase robustness to catastrophic forgetting \cite{Gallardo2021,Mehta2021} by enabling low-level feature reuse, as we will see in \cref{feature-reuse}. 

\paragraph{Optimization Strategy} We train the networks using SGD as an optimizer with momentum of 0.9 and weight decay of $3 \times 10^{-3}$. We use a polynomial learning rate schedule with power 0.9, and start CS training with a 0.07 learning rate and use a batch size 8. We use the same learning rate policy for ACDC subset training, starting with $5 \times 10^{-3}$. CS and ACDC subsets are trained for 200 and 150 epochs, respectively. We crop the images to $512 \times 1024$ and utilize random horizontal flipping and scaling during training. 
Unless stated otherwise, we do not use any other augmentations. During testing we do not use any scaling or cropping.

\paragraph{Evaluation Metrics:} We evaluate the performance of each model on the validation set of each dataset using the mean intersection-over-union (mIoU). We denote the mIoU of the model trained on all tasks up to $p$ and evaluated on task $q$ as $\text{mIoU}_{p, q}$. So that the zero-shot performance of a model trained on task $p=0$ and evaluated on $q=1$ is denoted as $\text{mIoU}_{0, 1}$. Furthermore, we report \textit{average learning accuracy} and \textit{forgetting}, that measure the learning capability and the severity of forgetting \cite{archmatters}. 
\begin{equation}
    \text{average learning acc.} = \frac{\text{mIoU}_{0, 0} + \text{mIoU}_{1, 1}}{2}
\end{equation}
\begin{equation}
    \text{forgetting} = \text{mIoU}_{0, 0} - \text{mIoU}_{1, 0}
\end{equation}
\subsection{Activation Drift after Incremental Adaptation}
First, we study the overall activation drift that arises in the CNN models when naively adapting to the different adverse weather conditions. 
Therefore, we first train DeepLabV3+ on CS and subsequently fine-tune it individually on each of the ACDC subsets. 
The results are displayed in \cref{tab:finetune}. 
As one would expect, the zero-shot performance on the adverse weather conditions is good for visually similar conditions such as \textit{Fog} and \textit{Rain} and significantly worse for \textit{Snow} and \textit{Night}, with \textit{Night} being the worst with only 10.4\% $\text{mIoU}_{0,1}$. This can be explained by the apparent differences between the domains, so that the day-to-night shift and snow-covered landscape represents a bigger shift than the wet environment or the foggy conditions~\cite{acdc}.
However, after fine-tuning on the adverse subsets, we notice that the better zero-shot performance on \textit{Rain} and \textit{Fog} does not indicate less forgetting compared to \textit{Snow} and \textit{Night}. Most strikingly, forgetting is the lowest after adapting to \textit{Night}. 
To determine which layers are most affected by the internal activation drift, we measure the activation drift for each layer between the model before and after learning the second task with layer stitching.
We use the setup explained in \cref{sec:layer_stitch}. 
The mIoU relative to the initial performance on the first task is shown in \cref{fig:layer_stitch_results}. 
We observe that activation drift is mainly affecting the network's early layers, which is in contrast to class-incremental learning settings, where early layers remain stable \cite{Kalb2022,davari2021probing,ramasesh2021anatomy}.
Specifically, the low-level features of the models tuned to \textit{Fog} and \textit{Snow} cannot be reused by the initial model, indicating that the shallow layers of the network have changed significantly. 
However, after the initial drop of relative mIoU in \cref{tab:finetune} we see a substantial increase in mIoU after \textit{layer1.0}, which indicates later features are in fact reused by the Cityscapes model. 
We hypothesize that at that point, features are more abstract and therefore more useful for the model trained on CS. 
After \textit{layer2.0} we observe a steady decrease in relative mIoU until the decoder layers.
We note that once the representations are shifted, subsequent layers are unlikely to regain similarity as their representations are based on the output of the previous layer.
The changing image distribution is most likely to blame for the initial feature disparity.
Thus, we analyze domain image distribution at the pixel level in the next section.

\subsection{Analysis of image statistics}
\begin{figure}
\centering
\includegraphics[width=\columnwidth]{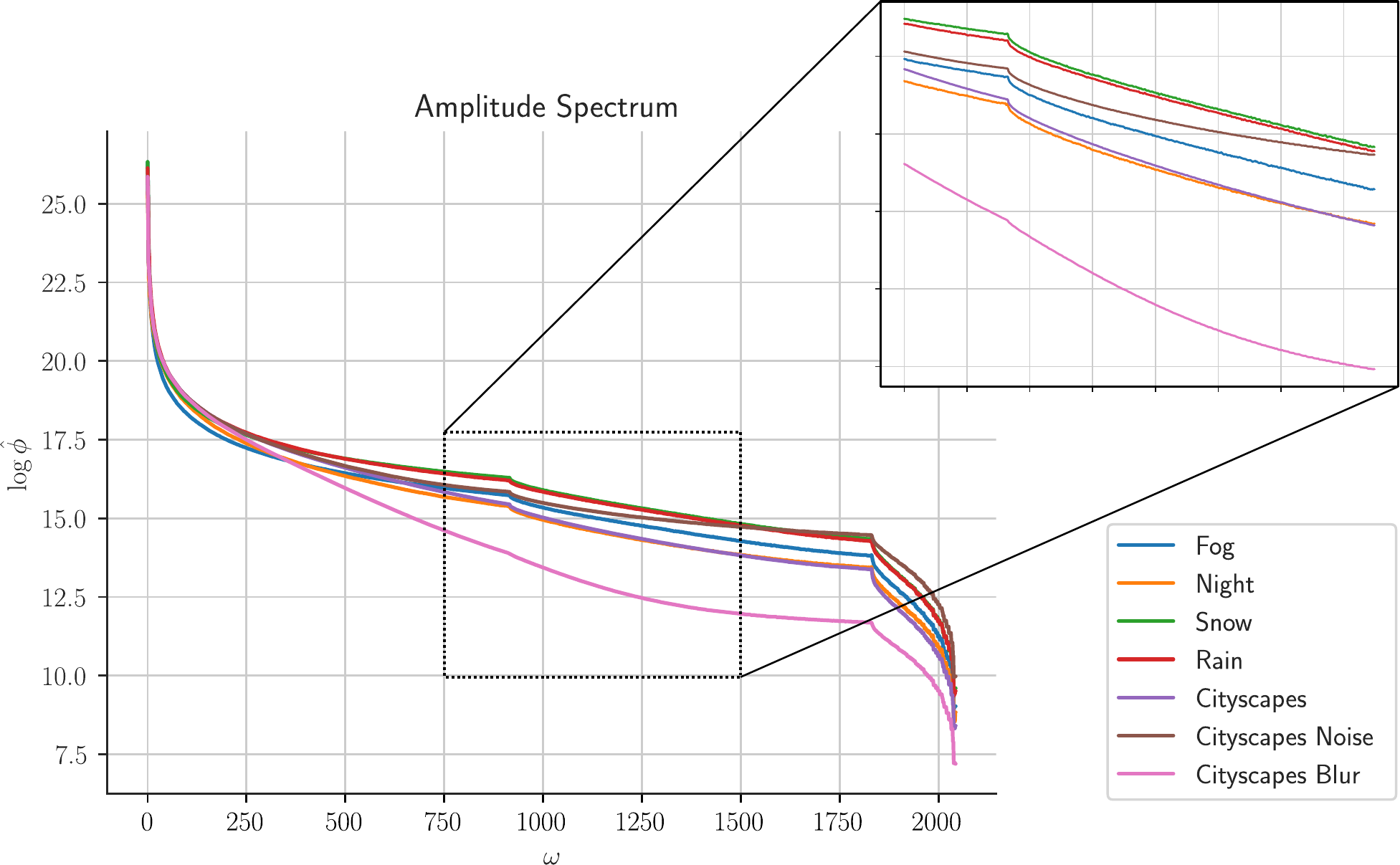}
\caption{Amplitude Spectra of Cityscapes, augmented CS and the ACDC subsets. In the frequency domain Cityscapes is much more similar to \textit{Night} than to any other of the ACDC subsets, specifically in the high-frequent components of the images. Blur is efficiently removing of high frequency components.}
\label{fig:fft1}
\end{figure}

To interpret the low-level feature change within the network, we compare the image statistics of each domain to explain the significant representation changes in the early layers.
Therefore, we compare the domains by their corresponding pixel mean and standard deviation for each HSV-channel in \cref{tab:color_hsv}. 
We observe that there is a substantial difference between the domains, specifically \textit{Rain}, \textit{Snow} and \textit{Fog} being notably brighter than Cityscapes.
Furthermore, as the capability of CNNs to generalize can be adversely affected by exploiting mid- to high-frequency components of images \cite{Wang_2020_CVPR, highfreq}, we also analyze the amplitude spectra in the frequency domain of the different training tasks in \cref{fig:fft1}. 
We can see that the domains are similar in low- to mid-frequency ranges, but \textit{Snow} and \textit{Rain} contain much more high-frequency components. This could lead to overfitting to high-frequency features when the model is trained on these domains and magnify forgetting.

\begin{table}
\centering
\resizebox{0.64\columnwidth}{!}{%
\begin{tabular}{l|ccc|ccc}
\multirow{2}{*}{Dataset} & \multicolumn{3}{c|}{Mean} & \multicolumn{3}{c}{Standard Deviation} \\ \cline{2-7} 
 & Hue & Saturation & Value & Hue & Saturation & Value \\ \hline
Rain & 86 & 32 & 110 & 62 & 35 & 78 \\
Snow & 104 & 19 & 132 & 55 & 21 & 62 \\
Night & 64 & 122 & 60 & 66 & 64 & 46 \\
Fog & 93 & 20 & 131 & 60 & 23 & 64 \\
Cityscapes & 59 & 49 & 83 & 18 & 22 & 49
\end{tabular}%
}
\caption{The Mean and standard deviations for the HSV-channels of each subset. There is a severe color shift between the domains in overall brightness of the images.} \label{tab:color_hsv}
\end{table}

\begin{table}
\centering
\resizebox{0.8\columnwidth}{!}{%
\begin{tabular}{@{}l|cc|ccc}
\toprule
 & \multicolumn{2}{c|}{\textbf{Task 1}} & \multicolumn{3}{c}{\textbf{Task 2}} \\
\textbf{Task 2} & $\text{mIoU}_{CS}$ & $\text{mIoU}_{T2}$ & $\text{mIoU}_{CS}$ & $\text{mIoU}_{T2}$ & Forgetting \\ \midrule
Rain & \cellcolor[HTML]{63BE7B}72.0 & \cellcolor[HTML]{AAD380}30.4 & \cellcolor[HTML]{F8696B}38.8 & \cellcolor[HTML]{FED880}57.7 & \cellcolor[HTML]{F8696B}33.2 \\
Night & \cellcolor[HTML]{63BE7B}72.0 & \cellcolor[HTML]{F8696B}10.5 & \cellcolor[HTML]{63BE7B}45.9 & \cellcolor[HTML]{F8696B}43.6 & \cellcolor[HTML]{63BE7B}26.1 \\
Snow & \cellcolor[HTML]{63BE7B}72.0 & \cellcolor[HTML]{FDCD7E}23.1 & \cellcolor[HTML]{FDCF7E}42.2 & \cellcolor[HTML]{D8E082}62.3 & \cellcolor[HTML]{FED07F}29.8 \\
Fog & \cellcolor[HTML]{63BE7B}72.0 & \cellcolor[HTML]{63BE7B}33.4 & \cellcolor[HTML]{CDDD82}44.0 & \cellcolor[HTML]{63BE7B}69.0 & \cellcolor[HTML]{CCDC81}28.0 \\ \bottomrule
\end{tabular}%
}
\caption{Results on \textit{CS} $\rightarrow$ \textit{ACDC} in mIoU (\%) for each subset of ACDC. While the zero-shot performance for \textit{Night} is the worst, after the fine-tuning to \textit{Night}, it is least affected by forgetting. }
\label{tab:finetune}
\end{table}

\subsection{Adjusting Low-Level Features} \label{feature-reuse} 
\begin{figure*}
\centering
\includegraphics[width=\textwidth]{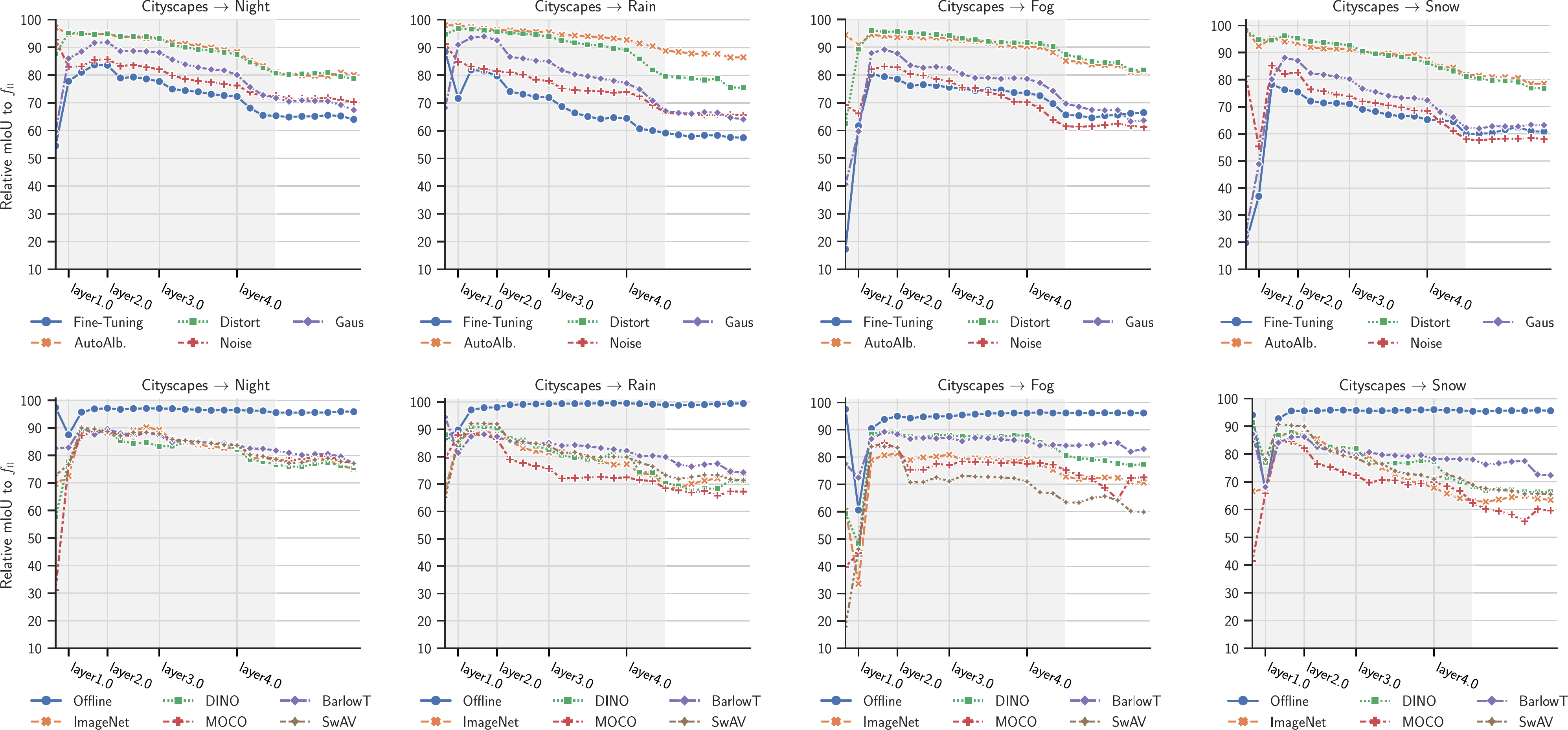}
\caption{Activation drift between $f_1$ to $f_0$ measured by relative mIoU on the first task of the model stitched together at specific layers (horizontal axis). The layers of the encoder are marked in the gray area, the decoder layers in the white area. The activations up until \textit{layer1.0} undergo drastic changes, specifically for \textit{Snow} and \textit{Fog}. After \textit{layer1.0} the activations can again be reused by $f_0$ leading to an mIoU increase. However, throughout the remaining encoder layers of the network the activations of $f_1$ further deviate from $f_0$.}
\label{fig:layer_stitch_pre}
\end{figure*}

Previous work suggests that low-level feature reuse is important for successful transfer learning \cite{NEURIPS2020_0607f4c7} and that pre-training can  mitigate forgetting for class-incremental classification tasks \cite{Gallardo2021,Mehta2021}. 
In a set of experiments, we show that low-level feature reuse is not only important for successful transfer of knowledge to downstream tasks, but is also vital to reduce catastrophic forgetting. 
Therefore, we investigate different pre-training and augmentation protocols to initialize the model $f_0$ on CS. 
We then fine-tune the model $f_0$ on the different ACDC subsets without any augmentations or any continual learning algorithm. 
Thus, we can examine how different methods improve the model's feature reuse and reduce forgetting.
To study the impact on pre-trained backbones, we use ResNet50 weights trained on ImageNet1k either fully-supervised or using the self-supervised methods DINO \cite{caron2021emerging}, MoCo v3 \cite{chen2021mocov3}, SwAV~\cite{caron2020unsupervised} and BarlowTwins \cite{zbontar2021barlow}.
For the augmentation experiments, we use the following strategies:
\begin{itemize}
    \setlength{\itemsep}{1pt}
    \setlength{\parskip}{0pt}
    \setlength{\parsep}{0pt}
    \setlength{\partopsep}{0pt}
    \item Using AutoAlbument (AutoAlbum) \cite{info11020125}, to learn an image augmentation policy from the CS dataset using Faster AutoAugment \cite{fasteraugment}.
    \item Learning color-invariant features by intensive Color Jittering and randomly re\-arrang\-ing input image channels. We denote this combination as \textit{Distortion} (Distort).
    \item Learning features tuned for mid- to low-frequencies by Gaussian blurring or adding Gaussian noise to remove high-frequency information. Changes to the spectrum are displayed in \cref{fig:fft1}
\end{itemize}
We also add an offline pre-trained model that is trained jointly on the CS and ACDC subsets and then fine-tuned on the target task. 
We use offline pre-training to infer an upper bound on a model's feature reuse, as the model should have learned features that are the joint optimum on both tasks.
The results are displayed in \cref{tab:pre-train}. 
The overall trend we see is that both pre-training and augmentations during training on CS lead to better transfer to the subsequent tasks and reduced forgetting on CS. 
However, while pre-training improves the transfer to the new tasks, it only moderately improves zero-shot capabilities when compared to the model without pre-training. 
Color-based augmentation and AutoAlbum improve zero-shot performance but perform worse on the ACDC tasks than the pre-trained models, indicating that better zero-shot performance does not always lead to better transfer performance.
Still, these augmentations are the most effective at mitigating forgetting for all tasks.
Augmentations based on removing high-frequency components reduce forgetting less significantly or, in the case of \textit{Fog}, lead to a further decrease in mIoU on the previous task. 
We note that for the domains \textit{Rain} and \textit{Snow} that contain more high-frequency components in their images than CS, adding noise and blurring is more effective than it is for \textit{Fog} and \textit{Night}.
So far, the results indicate two things: 1) pre-trained features are less susceptible to forgetting and lead to a better transfer to future tasks, 2) augmentations significantly improve generalization and produce more general features in the early layers. \\
We also investigate these findings using layer stitching and display the plots in \cref{fig:layer_stitch_pre}. 
The pre-trained models (bottom row) have a lower initial drop in similarity than randomly initialized models and remain higher throughout later layers.
Noticeable is that even the offline pre-trained model is affected by a severe drop in similarity for \textit{Snow}, \textit{Fog} and also a moderate drop in similarity for \textit{Rain} and \textit{Fog}. 
We will later confirm that this is largely due to the biased population mean and standard deviation of the BN layers.
Most surprising is that the models that used color augmentation during training are not affected by this initial drop, even though the optimization process on task 2 is the same for the offline and pre-trained models, so they should be affected by the same change in population statistics.
The fact that this does not occur indicates that when training with augmentations, the first convolutional layers extract features that are more domain-invariant than the features of the pre-trained models, resulting in BN layer population statistics that are less biased to the previous task \cite{pham2021continual}. We validate this claim in the next section.

\begin{table*}
\resizebox{\textwidth}{!}{%
\begin{tabular}{@{}cl|c|ccc|ccc|ccc|ccc}
\toprule
\multicolumn{2}{c|}{} & \textbf{Cityscapes} & \multicolumn{3}{c|}{\textbf{Night}} & \multicolumn{3}{c|}{\textbf{Rain}} & \multicolumn{3}{c|}{\textbf{Fog}} & \multicolumn{3}{c}{\textbf{Snow}} \\
\multicolumn{2}{c|}{} & \textit{Test} & \textit{Zero} & \textit{Test} & \textit{} & \textit{Zero} & \textit{Test} & \textit{} & \textit{Zero} & \textit{Test} & \textit{} & \textit{Zero} & \textit{Test} & \textit{} \\
\multicolumn{2}{c|}{\multirow{-3}{*}{\textbf{Method}}} & \textit{mIoU} & \textit{Shot} & \textit{mIoU} & \textit{Forgetting} & \textit{Shot} & \textit{mIoU} & \textit{Forgetting} & \textit{Shot} & \textit{mIoU} & \textit{Forgetting} & \textit{Shot} & \textit{mIoU} & \textit{Forgetting} \\ \midrule
 & FT & \cellcolor[HTML]{FDC87D}72.0 & \cellcolor[HTML]{FCBC7B}10.5 & \cellcolor[HTML]{F8696B}43.6 & \cellcolor[HTML]{F8696B}26.1 & \cellcolor[HTML]{FEE082}30.4 & \cellcolor[HTML]{F8696B}57.7 & \cellcolor[HTML]{F8696B}33.2 & \cellcolor[HTML]{FCC57C}33.4 & \cellcolor[HTML]{FEE582}69.0 & \cellcolor[HTML]{FCA978}28.0 & \cellcolor[HTML]{FCB97A}23.1 & \cellcolor[HTML]{F8696B}62.3 & \cellcolor[HTML]{F9766E}33.2 \\
 & AutoAlb. & \cellcolor[HTML]{FDCF7E}72.2 & \cellcolor[HTML]{63BE7B}24.2 & \cellcolor[HTML]{FFEB84}47.3 & \cellcolor[HTML]{63BE7B}15.1 & \cellcolor[HTML]{63BE7B}42.8 & \cellcolor[HTML]{FBAE78}59.4 & \cellcolor[HTML]{63BE7B}10.7 & \cellcolor[HTML]{63BE7B}50.1 & \cellcolor[HTML]{FDCA7D}68.2 & \cellcolor[HTML]{68BF7B}14.7 & \cellcolor[HTML]{63BE7B}37.2 & \cellcolor[HTML]{FBAF78}63.8 & \cellcolor[HTML]{63BE7B}10.7 \\
 & Distort & \cellcolor[HTML]{FCBE7B}71.7 & \cellcolor[HTML]{A0D07F}19.8 & \cellcolor[HTML]{FDCE7E}46.5 & \cellcolor[HTML]{99CD7E}16.2 & \cellcolor[HTML]{96CD7E}38.9 & \cellcolor[HTML]{FFEB84}60.9 & \cellcolor[HTML]{C4DA80}19.0 & \cellcolor[HTML]{92CC7E}46.5 & \cellcolor[HTML]{FDCD7E}68.3 & \cellcolor[HTML]{71C27B}15.2 & \cellcolor[HTML]{ABD380}32.8 & \cellcolor[HTML]{F8696B}62.3 & \cellcolor[HTML]{B0D47F}19.0 \\
 & Gaus & \cellcolor[HTML]{F8696B}69.1 & \cellcolor[HTML]{F98C71}8.1 & \cellcolor[HTML]{FDC77D}46.3 & \cellcolor[HTML]{FB9D75}23.0 & \cellcolor[HTML]{F8696B}26.9 & \cellcolor[HTML]{FDC67C}60.0 & \cellcolor[HTML]{FDC57D}26.7 & \cellcolor[HTML]{FA9172}26.9 & \cellcolor[HTML]{F8696B}65.4 & \cellcolor[HTML]{FCA577}28.3 & \cellcolor[HTML]{F8696B}15.8 & \cellcolor[HTML]{FEDA80}64.7 & \cellcolor[HTML]{F6E883}26.5 \\
\multirow{-5}{*}{{\rotatebox[origin=c]{90}{\parbox[c]{1cm}{\centering Augment.}}}} & Noise & \cellcolor[HTML]{F9806F}69.8 & \cellcolor[HTML]{FBAA77}9.6 & \cellcolor[HTML]{FDD57F}46.7 & \cellcolor[HTML]{FDB97B}21.3 & \cellcolor[HTML]{F98770}27.8 & \cellcolor[HTML]{FEDE81}60.6 & \cellcolor[HTML]{FFDA81}25.2 & \cellcolor[HTML]{FA9673}27.5 & \cellcolor[HTML]{F8E984}69.3 & \cellcolor[HTML]{FA8C72}30.2 & \cellcolor[HTML]{FBA576}21.3 & \cellcolor[HTML]{FA9373}63.2 & \cellcolor[HTML]{FCB37A}30.2 \\ \hline
 & ImageNet & \cellcolor[HTML]{D8E082}73.9 & \cellcolor[HTML]{F8696B}6.3 & \cellcolor[HTML]{F2E884}47.5 & \cellcolor[HTML]{FFDD82}19.1 & \cellcolor[HTML]{F8766D}27.3 & \cellcolor[HTML]{FFEB84}60.9 & \cellcolor[HTML]{EDE683}22.5 & \cellcolor[HTML]{F8696B}21.8 & \cellcolor[HTML]{FEDE81}68.8 & \cellcolor[HTML]{FFE283}23.8 & \cellcolor[HTML]{FDD680}25.8 & \cellcolor[HTML]{A4D17F}66.2 & \cellcolor[HTML]{FED07F}28.8 \\
 & MOCO & \cellcolor[HTML]{9BCF7F}75.2 & \cellcolor[HTML]{EFE784}14.0 & \cellcolor[HTML]{BFD981}48.3 & \cellcolor[HTML]{CFDD81}17.3 & \cellcolor[HTML]{E7E483}32.6 & \cellcolor[HTML]{ABD380}63.5 & \cellcolor[HTML]{FDC37D}26.8 & \cellcolor[HTML]{CEDD82}41.9 & \cellcolor[HTML]{68C07C}72.3 & \cellcolor[HTML]{F1E783}22.3 & \cellcolor[HTML]{D4DF82}30.3 & \cellcolor[HTML]{9CCF7F}66.3 & \cellcolor[HTML]{F8696B}33.8 \\
 & DINO & \cellcolor[HTML]{A5D17F}75.0 & \cellcolor[HTML]{FDD27F}11.6 & \cellcolor[HTML]{63BE7B}49.7 & \cellcolor[HTML]{FFE483}18.7 & \cellcolor[HTML]{FA9573}28.2 & \cellcolor[HTML]{8ECB7E}64.4 & \cellcolor[HTML]{F8E983}23.4 & \cellcolor[HTML]{FDD47F}35.2 & \cellcolor[HTML]{63BE7B}72.4 & \cellcolor[HTML]{C0D980}19.6 & \cellcolor[HTML]{FEE783}27.3 & \cellcolor[HTML]{63BE7B}67.0 & \cellcolor[HTML]{FBEA83}27.1 \\
 & BarlowT. & \cellcolor[HTML]{D8E082}73.9 & \cellcolor[HTML]{EBE683}14.3 & \cellcolor[HTML]{FFEB84}47.3 & \cellcolor[HTML]{C6DA80}17.1 & \cellcolor[HTML]{CBDC81}34.8 & \cellcolor[HTML]{63BE7B}65.7 & \cellcolor[HTML]{EDE683}22.5 & \cellcolor[HTML]{D2DE82}41.6 & \cellcolor[HTML]{93CC7E}71.4 & \cellcolor[HTML]{63BE7B}14.4 & \cellcolor[HTML]{BCD881}31.8 & \cellcolor[HTML]{E4E383}65.4 & \cellcolor[HTML]{CCDC81}22.0 \\
 & SwAV & \cellcolor[HTML]{63BE7B}76.4 & \cellcolor[HTML]{D6E082}15.8 & \cellcolor[HTML]{CBDC81}48.1 & \cellcolor[HTML]{E8E482}17.8 & \cellcolor[HTML]{FCEA84}31.0 & \cellcolor[HTML]{CFDD82}62.4 & \cellcolor[HTML]{FFE483}24.5 & \cellcolor[HTML]{DBE182}40.9 & \cellcolor[HTML]{85C87D}71.7 & \cellcolor[HTML]{F8696B}32.8 & \cellcolor[HTML]{FAEA84}28.0 & \cellcolor[HTML]{93CC7E}66.4 & \cellcolor[HTML]{FFE483}27.8 \\
\multirow{-6}{*}{{\rotatebox[origin=c]{90}{\parbox[c]{1cm}{\centering Pre-Training}}}} & Offline & - & 46.1 & 47.4 & 3.2 & 59.0 & 59.1 & 0.5 & 67.5 & 66.9 & 3.2 & 62.4 & 62.9 & 3.3 \\ \bottomrule
\end{tabular}%
}
\caption{Results on \textit{CS} $\rightarrow$ \textit{ACDC} in mIoU (\%) for each subset of ACDC: \textit{Night}, \textit{Rain}, \textit{Snow} and \textit{Fog} using different pre-training and augmentation strategies (Augment.). While pre-training significantly improves learning accuracy, it does not improve zero-shot performance, but still gives moderate improvements in reducing forgetting. Augmentations lead to slightly improved performance on the target set, improved zero-shot performance and significantly reduced forgetting for all weather conditions. }
\label{tab:pre-train}
\end{table*}

\subsection{Impact of Batch Normalization on Forgetting}
\begin{figure}
\centering
\includegraphics[width=\columnwidth]{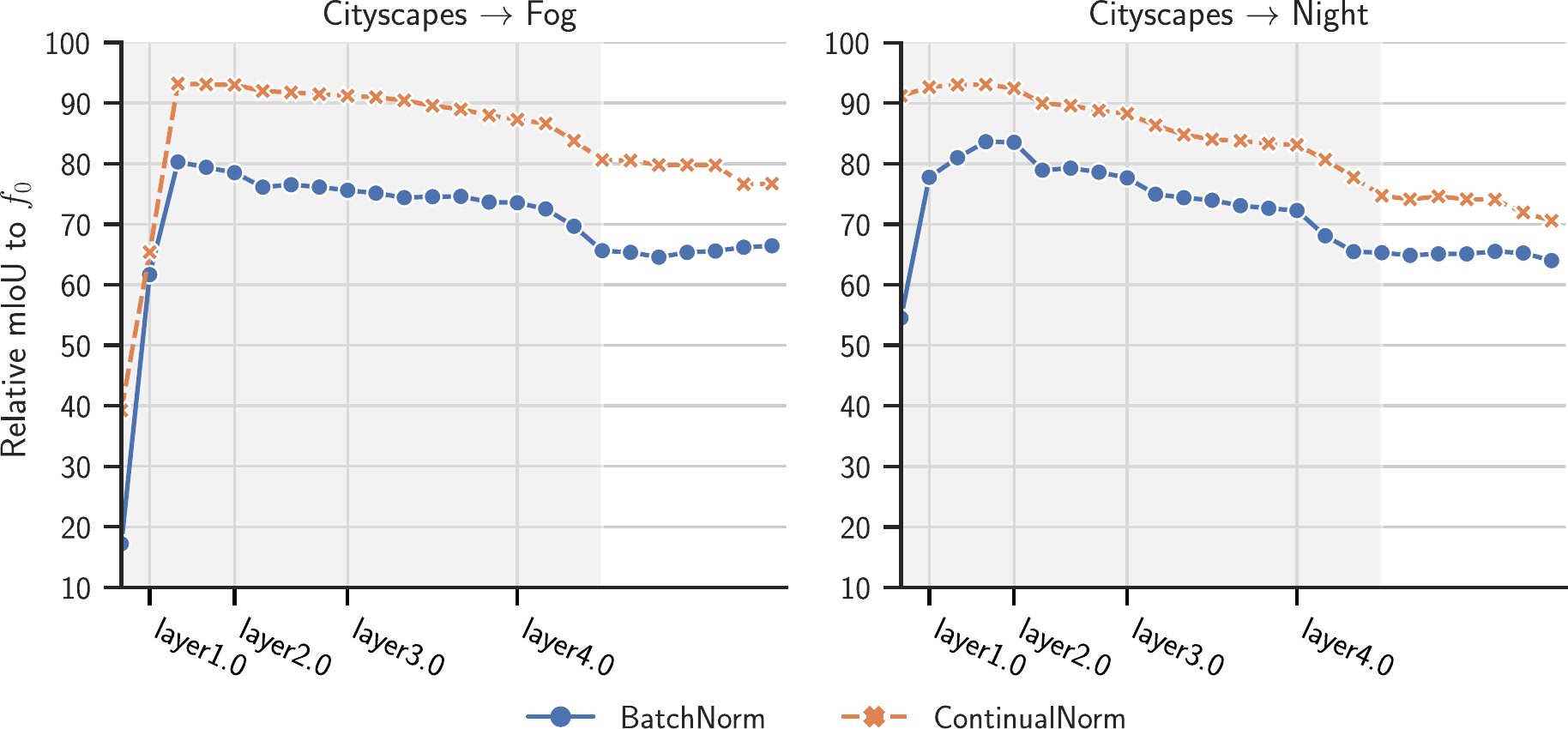}\hfill
\caption{Comparing the activation drift between models trained with Batch Normalization and Continual Normalization on \textit{Night} and \textit{Fog}. Continual Normalization effectively reduces forgetting by mitigating the biased population in statistics in early BN layers.}
\label{fig:cnvsbn}
\end{figure}

The results in \cref{feature-reuse} suggest that changing BN population statistics are a major cause of early layer representation changes. 
To verify this, we re-estimate the BN Layer population statistics on the combined dataset of CS and the specific ACDC subset without changing any parameters.\footnote{This is achieved by doing a forward pass over the dataset while the BN Layers update their population statistics}
We then evaluate the model on the CS dataset and report the re-estimated $\text{mIoU}_R$ and increased performance as $\Delta \text{mIoU}$ in \cref{tab:bnre_estimate}.
Most methods benefit significantly from re-estimation of population statistics, with the Fine-Tuning (FT) model benefiting the most.
Furthermore, we observe that pre-trained models improve only moderately compared to FT, meaning that they are less influenced by biased population statistics.
Finally, we discover that models trained with augmentation only slightly improve after BN re-estimation, and even decrease in the \textit{CS} $\rightarrow$ \textit{Fog} setting. 
We explain this effect by the invariance to low-level properties of the images such as hue, saturation and brightness that the first CNN layer of the models trained with augmentations has to learn in order to cope with the augmentation scheme. 
Thus, extracted features are invariant to these properties and subsequent BN layers are less affected by the distribution shift.
The fact that only re-estimating BN layers without changing the initial layers leads to such a significant improvement, demonstrates that the adjusted population statistics can normalize the variance between the domains.\footnote{In Appendix C, we investigate which BN layers are affected.}
Overall, this means that BN is a major contributor to forgetting in the domain-incremental setting, but forgetting is also precipitated by low-level features that are tuned to their specific domain, which lead to a major change in population statistics in the BN layers.
We validate these findings by exchanging all BN layers with Continual Normalization layers \cite{pham2021continual}. 
The results are shown in \cref{fig:cnvsbn} and \cref{tab:cnvsbn}.
We see that CN greatly reduces forgetting on CS, while we still observe a similar discrepancy in low-level features as for the models with BN. 
However, due to the combination of Group and Batch Normalization the changing low-level features are normalized across the channel dimensions before affecting the population statistics of BN.

\begin{table}
\resizebox{\columnwidth}{!}{%
\begin{tabular}{@{}c|c|ccc|ccc}
\toprule
\multirow{3}{*}{\textbf{Normalization}} & \textbf{CS} & \multicolumn{3}{c|}{\textbf{Night}} & \multicolumn{3}{c}{\textbf{Fog}} \\
 & \textit{Test} & \textit{Zero} & \textit{Test} & \textit{Average} & \textit{Zero} & \textit{Test} & \textit{Average} \\
 & \textit{mIoU} & \textit{Shot} & \textit{mIoU} & \textit{Forgetting} & \textit{Shot} & \textit{mIoU} & \textit{Forgetting} \\ \midrule
\multicolumn{1}{l|}{BatchNorm} & 72.0 & 10.5 & 43.6 & 26.1 & 33.4 & 69.0 & 28.0 \\
\multicolumn{1}{l|}{Cont. Norm} & 71.2 & 10.3 & 44.2 & 21.2 & 34.4 & 67.1 & 19.1 \\ \bottomrule
\end{tabular}%
}
\caption{Results on \textit{CS} $\rightarrow$ \textit{ACDC} in mIoU (\%) with Batch- and Continual Normalization. By reducing the biased population statistics in early BN layers, CN effectively reduces forgetting.}
\label{tab:cnvsbn}
\end{table}

\begin{table}
\resizebox{\columnwidth}{!}{%
\begin{tabular}{@{}l|cc|cc|cc|cc}
\toprule
\multicolumn{1}{c|}{} & \multicolumn{2}{c|}{\textbf{Night}} & \multicolumn{2}{c|}{\textbf{Rain}} & \multicolumn{2}{c|}{\textbf{Fog}} & \multicolumn{2}{c}{\textbf{Snow}} \\
\multicolumn{1}{c|}{\multirow{-2}{*}{\textbf{Method}}} & \textbf{\textit{$\text{mIoU}_R \uparrow$}} & \textbf{\textit{$\Delta \text{mIoU}$}} & \textbf{\textit{$\text{mIoU}_R \uparrow$}} & \textbf{\textit{$\Delta \text{mIoU}$}} & \textbf{\textit{$\text{mIoU}_R \uparrow$}} & \textbf{\textit{$\Delta \text{mIoU}$}} & \textbf{\textit{$\text{mIoU}_R \uparrow$}} & \textbf{\textit{$\Delta \text{mIoU}$}} \\ \midrule
FT & 58.6 & \cellcolor[HTML]{F8696B}12.7 & 58.2 & \cellcolor[HTML]{F8696B}19.4 & 49.7 & \cellcolor[HTML]{FCA377}5.8 & 51.3 & \cellcolor[HTML]{FDB97B}9.1 \\
AutoAlb. & 59.8 & \cellcolor[HTML]{63BE7B}2.7 & 62.0 & \cellcolor[HTML]{72C27B}0.5 & 54.4 & \cellcolor[HTML]{83C77C}-3.1 & 53.9 & \cellcolor[HTML]{63BE7B}-2.1 \\
Distort & 59.2 & \cellcolor[HTML]{8CC97D}3.7 & 59.1 & \cellcolor[HTML]{CADB80}6.5 & 52.0 & \cellcolor[HTML]{63BE7B}-4.5 & 54.0 & \cellcolor[HTML]{93CB7D}0.0 \\
ImageNet & 61.7 & \cellcolor[HTML]{FED981}7.4 & 59.6 & \cellcolor[HTML]{E1E282}8.1 & 51.7 & \cellcolor[HTML]{FDEA83}2.1 & 54.7 & \cellcolor[HTML]{FCB279}9.7 \\
MOCO & 63.5 & \cellcolor[HTML]{DAE081}5.6 & 62.5 & \cellcolor[HTML]{FDB57A}14.0 & 55.1 & \cellcolor[HTML]{FFEB84}2.2 & 57.4 & \cellcolor[HTML]{F8696B}16,0 \\
DINO & 64.0 & \cellcolor[HTML]{FED280}7.7 & 63.8 & \cellcolor[HTML]{FECF7F}12.2 & 57.8 & \cellcolor[HTML]{FFE784}2.4 & 61.9 & \cellcolor[HTML]{FA8170}14.0 \\
BarlowT & 65.4 & \cellcolor[HTML]{FDBF7C}8.6 & 65.2 & \cellcolor[HTML]{FDB77A}13.9 & 60.6 & \cellcolor[HTML]{F8696B}8.7 & 59.4 & \cellcolor[HTML]{93CB7D}0.0 \\
Img+Dis & 62.5 & \cellcolor[HTML]{A4D07E}4.3 & 63.0 & \cellcolor[HTML]{63BE7B}-0.6 & 55.3 & \cellcolor[HTML]{65BE7B}-4.4 & 59.0 & \cellcolor[HTML]{9ACD7E}0.3 \\
Offline & 68.3 & -0.2 & 70.3 & -0.8 & 68.8 & -0.4 & 69.2 & 0.4 \\ \bottomrule
\end{tabular}%
}
\caption{Performance in mIoU [\%] on \textit{CS} of the model $f_1$ after re-estimating all BN layer population statistics. FT is most affected by changing population statistics, while models trained with augmentation are least affected.}
\label{tab:bnre_estimate}
\end{table}

\subsection{Combining our Findings}\label{ablate_inc}
\begin{figure}
\centering
\includegraphics[width=\columnwidth]{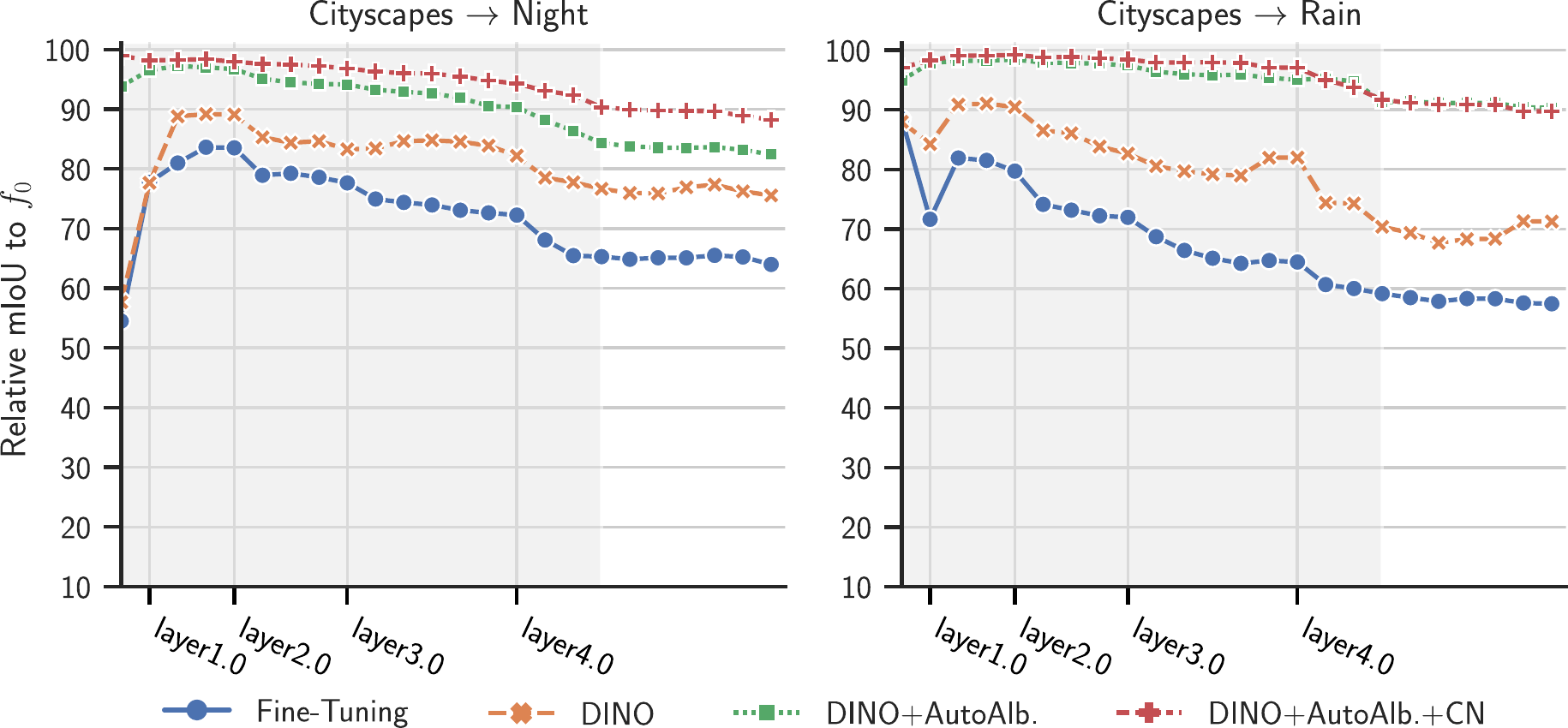}\hfill
\caption{Comparing the activation drift between models trained on \textit{Night} and \textit{Fog} with sequentially adding pre-training with DINO, AutoAlb. and CN.}
\label{fig:seq_addstitch}
\end{figure}

We run the same experiments making incremental changes to the training process in CS by sequentially adding pre-training with DINO, then AutoAlbum and replacing BN with CN layers. 
Results in \cref{tab:ablate_inc} demonstrate that these changes complement each other as they dramatically reduce forgetting on CS. 
This is apparent when comparing the layer stitching plots in \cref{fig:seq_addstitch}, where we see that pre-training with DINO alone increases feature reuse only after \textit{layer1.0} compared to fine-tuning. 
However, when combined with augmentations and CN, the representation drift before \textit{layer1.0} is significantly reduced as well. 
This indicates that pre-training and training with augmentations enable feature reuse at different layers of the network depending on the task at hand.
Therefore, when combining pre-training, CN and augmentation, we increase the feature reuse in all domains, as we see in \cref{fig:layer_stitch_results}, reducing forgetting without any continual learning algorithm.

\begin{table}
\centering
\resizebox{0.9\columnwidth}{!}{%
\begin{tabular}{@{}l|cc|cc|cc|cc}
\toprule
\multicolumn{1}{c|}{} & \multicolumn{2}{c|}{\textbf{Night}} & \multicolumn{2}{c|}{\textbf{Rain}} & \multicolumn{2}{c|}{\textbf{Fog}} & \multicolumn{2}{c}{\textbf{Snow}} \\
\multicolumn{1}{c|}{} & \textit{lrn.} & \textit{} & \textit{lrn.} & \textit{} & \textit{lrn.} & \textit{} & \textit{lrn.} & \textit{} \\
\multicolumn{1}{c|}{\multirow{-3}{*}{\textbf{Method}}} & \textit{acc.} & \textit{forg.} & \textit{acc.} & \textit{forg.} & \textit{acc.} & \textit{forg.} & \textit{acc.} & \textit{forg.} \\ \midrule
FT & \cellcolor[HTML]{F8696B}57.8 & \cellcolor[HTML]{F8696B}26.1 & \cellcolor[HTML]{F8696B}64.9 & \cellcolor[HTML]{F8696B}33.2 & \cellcolor[HTML]{F8696B}70.5 & \cellcolor[HTML]{F8696B}28.0 & \cellcolor[HTML]{F8696B}67.2 & \cellcolor[HTML]{F8696B}29.8 \\
+ DINO & \cellcolor[HTML]{63BE7B}62.3 & \cellcolor[HTML]{FECA7E}18.7 & \cellcolor[HTML]{FEDB81}69.7 & \cellcolor[HTML]{FCB37A}23.4 & \cellcolor[HTML]{FEE783}73.7 & \cellcolor[HTML]{FDC27D}19.6 & \cellcolor[HTML]{E0E383}71.0 & \cellcolor[HTML]{FA8E72}27.1 \\
+ AutoAlb. & \cellcolor[HTML]{80C77D}62.2 & \cellcolor[HTML]{C5DA80}13.5 & \cellcolor[HTML]{63BE7B}71.2 & \cellcolor[HTML]{66BF7B}8.2 & \cellcolor[HTML]{F6E984}73.9 & \cellcolor[HTML]{B5D57F}11.7 & \cellcolor[HTML]{63BE7B}71.2 & \cellcolor[HTML]{8DCA7D}13.1 \\
+ CN & \cellcolor[HTML]{FEDC81}61.3 & \cellcolor[HTML]{63BE7B}9.1 & \cellcolor[HTML]{88C97E}71.0 & \cellcolor[HTML]{63BE7B}8.0 & \cellcolor[HTML]{63BE7B}75.4 & \cellcolor[HTML]{63BE7B}7.3 & \cellcolor[HTML]{FEE983}70.9 & \cellcolor[HTML]{63BE7B}10.5 \\ \bottomrule
\end{tabular}%
}
\caption{Forgetting and Learning accuracy on \textit{CS} $\rightarrow$ \textit{ACDC} with incremental additions that increase the feature reuse, significantly reduces forgetting.}
\label{tab:ablate_inc}
\end{table}

\section{Ablation Studies}

\subsection{Comparison to Continual Learning Algorithms}\label{compareSOA} 
So far, no explicit continual learning strategies like regularization or replay were used. 
Therefore, we show in \cref{tab:ablate_inc_cl} that achieving low-level feature reuse outperforms regularization methods even when using the same initialization. 
However, our training regime is still outperformed by naive replay. 
While EWC significantly improves when combined with CN and AutoAlbum, we see that Replay does not benefit from these adjustments. 
For Replay only pre-training leads to a significant increase in learning accuracy and a minor reduction of forgetting.
A likely explanation is that the model is able to learn features that are invariant to domain differences due to the batch construction during replay, in which half of the mini-batch consists of replay samples from previous tasks.
This could explain why replay is reported to be sample efficient in domain incremental learning \cite{Kalb2021}.

\begin{table}
\centering
\resizebox{0.75\columnwidth}{!}{%
\begin{tabular}{@{}l|cc|cc}
\toprule
\multicolumn{1}{c|}{} & \multicolumn{2}{c|}{\textbf{Night}} & \multicolumn{2}{c}{\textbf{Rain}} \\
\multicolumn{1}{c|}{} & \textit{lrn.} & \textit{avg.} & \textit{lrn.} & \textit{avg.} \\
\multicolumn{1}{c|}{\multirow{-3}{*}{\textbf{Method}}} & \textit{acc.} & \textit{forg.} & \textit{acc.} & \textit{forg.} \\ \midrule
EWC & \cellcolor[HTML]{F8696B}52.8 & \cellcolor[HTML]{F8696B}18.0 & \cellcolor[HTML]{F8696B}61.4 & \cellcolor[HTML]{F8696B}21.2 \\
+ DINO & \cellcolor[HTML]{FEE182}58.1 & \cellcolor[HTML]{FDBD7C}10.0 & \cellcolor[HTML]{FCB679}66.0 & \cellcolor[HTML]{FDB87B}11.3 \\
+ AutoAlb + CN & \cellcolor[HTML]{FFEB84}58.5 & \cellcolor[HTML]{ABD27F}4.8 & \cellcolor[HTML]{FFEB84}69.1 & \cellcolor[HTML]{FFEB84}4.9 \\ \hline
Replay & \cellcolor[HTML]{FEE482}58.2 & \cellcolor[HTML]{FFEB84}5.6 & \cellcolor[HTML]{FA9874}64.2 & \cellcolor[HTML]{8FCA7D}2.9 \\
+ DINO & \cellcolor[HTML]{63BE7B}62.4 & \cellcolor[HTML]{6DC07B}4.2 & \cellcolor[HTML]{74C37C}70.8 & \cellcolor[HTML]{63BE7B}2.1 \\
+ AutoAlb + CN & \cellcolor[HTML]{7CC57D}61.8 & \cellcolor[HTML]{63BE7B}4.1 & \cellcolor[HTML]{BED981}69.9 & \cellcolor[HTML]{C1D980}3.8 \\ \hline
DINO + AutoAlb + CN & \cellcolor[HTML]{6FC27C}62.1 & \cellcolor[HTML]{FED380}7.9 & \cellcolor[HTML]{63BE7B}71.0 & \cellcolor[HTML]{FED380}8.0 \\ \bottomrule
\end{tabular}%
}
\caption{Forgetting and Learning accuracy on \textit{Night} and \textit{Rain} with EWC and Replay. Our training scheme outperforms EWC on this benchmark and can be added to existing CL methods to improve learning accuracy and reduce forgetting.}
\label{tab:ablate_inc_cl}
\end{table}

\subsection{Architectures}\label{segformer}
We confirm our findings by repeating our experiments on SegFormer~\cite{xie2021segformer}, DeepLabV3+, and ERFNet~\cite{erfnet}, all of which were pre-trained on ImageNet.
We choose DeepLabV3+ with a ResNet50 backbone and SegFormer-B2 as they have a similar number of parameters, so that the results are comparable. 
The much smaller ERFNet is known to be more susceptible to forgetting due to its size~\cite{Kalb2022}. 
The results in \cref{tab:arch_compare} show that transformer-based SegFormer is much less affected by catastrophic forgetting than its CNN counterpart, even without any augmentations. 
A likely explanation is that SegFormer learns more general features and is thus more robust to distribution changes. This would also explain why SegFormer-B2 does not improve as significantly as DeepLabV3+ with the addition of pixel-level augmentations, as shown \cref{tab:segformeraug}. 
However, it is unclear whether this is an inherent feature of the self-attention mechanism, a result of training recipes \cite{NEURIPS2021_e19347e1}, or of architectural choices such as using Layer- instead of Batch Normalization. We leave this for future studies.

\begin{table}
\resizebox{\columnwidth}{!}{%
\begin{tabular}{@{}l|cc|cc|cc|cc}
\toprule
\textbf{} & \multicolumn{2}{c|}{\textbf{Night}} & \multicolumn{2}{c|}{\textbf{Rain}} & \multicolumn{2}{c|}{\textbf{Fog}} & \multicolumn{2}{c}{\textbf{Snow}} \\
\multicolumn{1}{c|}{} & \multicolumn{1}{c|}{\textit{lrn.}} & \textit{} & \textit{lrn.} & \textit{} & \textit{lrn.} & \textit{} & \textit{lrn.} & \textit{} \\
\multicolumn{1}{c|}{\multirow{-2}{*}{Model}} & \multicolumn{1}{c|}{\textit{acc.}} & \textit{forg.} & \textit{acc.} & \textit{forg.} & \textit{acc.} & \textit{forg.} & \textit{acc.} & \textit{forg.} \\ \midrule
Segfromer-B2 \cite{NEURIPS2021_64f1f27b} & \multicolumn{1}{c|}{\cellcolor[HTML]{FFEB84}59.4} & \cellcolor[HTML]{63BE7B}16.2 & \cellcolor[HTML]{63BE7B}69.1 & \cellcolor[HTML]{63BE7B}11.1 & \cellcolor[HTML]{63BE7B}71.6 & \cellcolor[HTML]{63BE7B}11.8 & \cellcolor[HTML]{63BE7B}70.4 & \cellcolor[HTML]{63BE7B}14.3 \\
DeepLabv3+ \cite{Chen_2018_ECCV} & \multicolumn{1}{c|}{\cellcolor[HTML]{63BE7B}60.5} & \cellcolor[HTML]{FFEB84}19.1 & \cellcolor[HTML]{FFEB84}67.4 & \cellcolor[HTML]{FFEB84}22.5 & \cellcolor[HTML]{FFEB84}71.1 & \cellcolor[HTML]{FFEB84}23.8 & \cellcolor[HTML]{FFEB84}70 & \cellcolor[HTML]{FFEB84}28.8 \\
ERFNet \cite{erfnet} & \multicolumn{1}{c|}{\cellcolor[HTML]{F8696B}56.6} & \cellcolor[HTML]{F8696B}29 & \cellcolor[HTML]{F8696B}63.2 & \cellcolor[HTML]{F8696B}36.2 & \cellcolor[HTML]{F8696B}67.5 & \cellcolor[HTML]{F8696B}30.1 & \cellcolor[HTML]{F8696B}64.5 & \cellcolor[HTML]{F8696B}59.5 \\ \bottomrule
\end{tabular}%
}
\caption{Forgetting and Learning accuracy of different Segformer-B2, DeepLabv3+ and ERFNet trained on \textit{CS} $\rightarrow$ \textit{ACDC} in mIoU (\%) for each subset of ACDC.}
\label{tab:arch_compare}
\end{table}

\begin{table}
\resizebox{\columnwidth}{!}{%
\begin{tabular}{@{}l|cc|cc|cc|cc}
\toprule
\multicolumn{1}{c|}{} & \multicolumn{2}{c|}{\textbf{Night}} & \multicolumn{2}{c|}{\textbf{Rain}} & \multicolumn{2}{c|}{\textbf{Fog}} & \multicolumn{2}{c}{\textbf{Snow}} \\
\multicolumn{1}{c|}{} & \textit{lrn.} & \textit{} & \textit{lrn.} & \textit{} & \textit{lrn.} & \textit{} & \textit{lrn.} & \textit{} \\
\multicolumn{1}{c|}{\multirow{-3}{*}{Augment.}} & \textit{acc.} & \textit{forg.} & \textit{acc.} & \textit{forg.} & \textit{acc.} & \textit{forg.} & \textit{acc.} & \textit{forg.} \\ \midrule
 & \cellcolor[HTML]{F8696B}59.4 & \cellcolor[HTML]{F8696B}16.2 & \cellcolor[HTML]{63BE7B}69.1 & \cellcolor[HTML]{F8696B}11.1 & \cellcolor[HTML]{F8696B}71.6 & \cellcolor[HTML]{F8696B}11.8 & \cellcolor[HTML]{FFEB84}70.4 & \cellcolor[HTML]{F8696B}14.3 \\
Distort. & \cellcolor[HTML]{FFEB84}59.6 & \cellcolor[HTML]{63BE7B}14.0 & \cellcolor[HTML]{FFEB84}68.8 & \cellcolor[HTML]{FFEB84}9.6 & \cellcolor[HTML]{63BE7B}72.6 & \cellcolor[HTML]{FFEB84}10.1 & \cellcolor[HTML]{F8696B}70.3 & \cellcolor[HTML]{FFEB84}10.9 \\
AutoAlb. & \cellcolor[HTML]{63BE7B}59.7 & \cellcolor[HTML]{FFEB84}14.2 & \cellcolor[HTML]{F8696B}68.3 & \cellcolor[HTML]{63BE7B}8.5 & \cellcolor[HTML]{FFEB84}72.2 & \cellcolor[HTML]{63BE7B}7.9 & \cellcolor[HTML]{63BE7B}70.8 & \cellcolor[HTML]{63BE7B}10.5 \\ \bottomrule
\end{tabular}%
}
\caption{Forgetting and learning accuracy in mIoU (\%) of SegFormer-B2 with different augmentations SegFormer-B2 improves less than DeepLabV3+ using augmentations, suggesting it is less affected by color-dependent features. }
\label{tab:segformeraug}
\end{table}

\section{Conclusion}
Our study has shown that a major cause of catastrophic forgetting in domain-incremental learning is the shift of low-level representations, particularly in the first convolution layer. 
This shift affects the population statistics of subsequent BN layers and results in forgetting when adapting to new domains.
To address this problem, we investigated the use of various pre-training schemes and pixel-level augmentations to facilitate features in early layers that can be reused in upcoming tasks. 
Our experiments showed that these methods were effective in reducing representation shift, with pre-training stabilizing the first layers, and augmentations primarily stabilizing the representations after the first BN layer.
We hypothesize that training with augmentation strategies like \emph{Distortion} or \emph{AutoAlbum} encourages the model to learn features that are invariant to low-level image statistics such as hue, saturation and brightness that vary between the domains. 
So that during optimization on the new domain those features are not affected, leading to a significant reduction in forgetting.
Interestingly, we found that pre-trained models struggle to learn such features in the early layers, but they still reduce forgetting notably compared to randomly initialized models. 
Leading us to believe that pre-training on ImageNet leads to more generalized features throughout the network. 
In our experiments, self-supervised pre-training outperformed supervised ImageNet pre-training on all domains, which suggests that SSL pre-training might not only be a vital tool for classification \cite{Gallardo2021}, but also for continual semantic segmentation.\\
The effectiveness of these approaches varies across the domains but is consistent for different CNN architectures that use BN layers\footnote{Experiments on architectures are displayed in Appendix F}. 
Our findings are related to the research on spurious features in continual learning \cite{lesort2022continual}, as our training scheme is reducing the emergence of spurious features for the first task. 
Overall, we hope that our results highlight that an important component of continual learning can be found in methods that extract generalized features from the initial task, instead of only mitigating the effects of catastrophic forgetting during training on new data.\\
However, we note that even with improved low-level feature reuse, the model is still susceptible to catastrophic forgetting in later layers due to more abstract domain changes.
As a result, we consider low-level feature reuse in incremental learning to be an important component of continual learning, but not the sole solution.

\section*{ACKNOWLEDGMENT}
The research leading to these results is funded by the German Federal Ministry for Economic Affairs and Climate Action within the project “KI Delta Learning“ (Förderkennzeichen 19A19013T).

{\small
\bibliographystyle{ieee_fullname}
\bibliography{egbib}
}

\clearpage

\appendix
\section{Implementation Details}
\subsection{Models and Weights}
All of our experiments are conducted using PyTorch in combination with PyTorch Lightning. 
We use the PyTorch implementation of ERFNet provided by \cite{erfnet}, which can be found at: \href{https://github.com/Eromera/erfnet_pytorch}{github.com/Eromera/erfnet\_pytorch}, the DeepLabV3+ \cite{Chen_2018_ECCV} implementation from \textit{Segmentation Models PyTorch} \cite{Iakubovskii:2019} and the SegFormer-B2 \cite{xie2021segformer} implementation from HuggingFace Transformers \cite{wolf-etal-2020-transformers}. The weights for the pre-trained ResNet-50 \cite{He2015} backbones are taken from:
\begin{itemize}
	\setlength{\itemsep}{1pt}
	\setlength{\parskip}{0pt}
	\setlength{\parsep}{0pt}
	\setlength{\partopsep}{0pt}
	\item DINO \cite{caron2021emerging}: \href{https://github.com/facebookresearch/dino}{github.com/facebookresearch/dino}
	\item MoCo v3 \cite{chen2021mocov3}: \href{github.com/facebookresearch/moco-v3}{github.com/facebookresearch/moco-v3}
	\item BarlowTwins \cite{zbontar2021barlow}: \href{https://github.com/facebookresearch/barlowtwins}{github.com/facebookresearch/barlow twins}
	\item SwAV \cite{caron2020unsupervised}: \href{https://github.com/facebookresearch/swav}{github.com/facebookresearch/swav}
\end{itemize} 
The weights of ERFNet pre-trained with DINO and MoCo v3 can be found on \href{https://github.com/tobiaskalb/feature-reuse-css}{github.com/tobiaskalb/feature-reuse-css}.

\subsection{Hyperparameter Choice}
For each model we start by tuning the LR on Cityscapes. We ran experiments with LR $\in$ \{0.1, 0.05, 0.01, 0.005, 0.001, 0.0005\}. 
We test intermediate LRs between the best and second-best LR. 
For the pre-trained and augmentation models, we choose the same LR. 
We chose the parameters for FT and EWC using the Continual Hyperparameter Framework~\cite{delange2021clsurvey}.

\subsection{Augmentations}
For all our augmentations we utilize Albumentations \cite{info11020125}. The augmentation schemes and their specific configurations that were used in our experiments are shown in \cref{tab:augs}. The config of the AutoAlbum and further information on the transformation pipelines can be found at: \href{https://github.com/tobiaskalb/feature-reuse-css}{github.com/tobiaskalb/feature-reuse-css}. We chose the parameters of Distort to be similar to \textit{PhotometricDistortion} in \href{https://mmsegmentation.readthedocs.io/en/latest/api.html#mmseg.datasets.pipelines.PhotoMetricDistortion}{MMSegmentation}.

\begin{table}[h]
	\resizebox{\columnwidth}{!}{%
		\begin{tabular}{l|l}
			\textbf{Method} & \textbf{Albumentations Parameters} \\ \hline
			Distortion & ColorJitter(brightness=0.2, contrast=0.5, saturation=0.5, hue=0.2) \\
			& ChannelShuffle(p=0.5) \\ \hline
			Gaussian Blur & GaussianBlur(blur\_limit=(3, 5)) \\ \hline
			Gaussian Noise & GaussNoise(var\_limit=(30, 60) \\ \hline
			AutoAlbument & Augementation json config
		\end{tabular}%
	}
	\caption{Additional augmentations used in the experiments with there specified arguments and classes using Albumentations \cite{info11020125}. }
	\label{tab:augs}
\end{table}

\section{Amplitude Spectra of ACDC and Cityscapes}
\begin{figure}
	\resizebox{\columnwidth}{!}{%
		\begin{tabular}{ccc}
			\textbf{Cityscapes} & \textbf{Cityscapes Blur} & \textbf{Cityscapes Noise} \\
			\includegraphics[width=0.5\columnwidth]{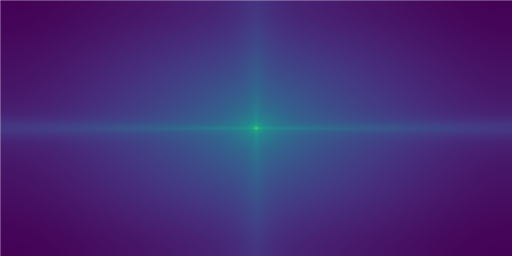} & 
			\includegraphics[width=0.5\columnwidth]{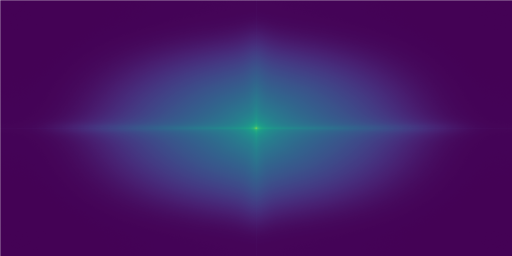}  &  
			\includegraphics[width=0.5\columnwidth]{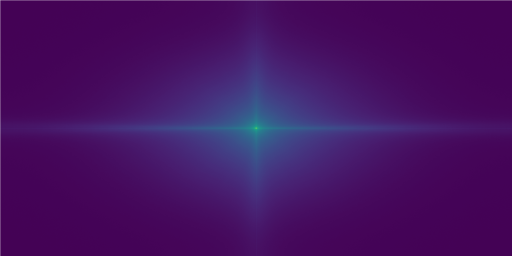} \\
			\textbf{Fog} & \textbf{Night} & \textbf{Snow} \\
			\includegraphics[width=0.5\columnwidth]{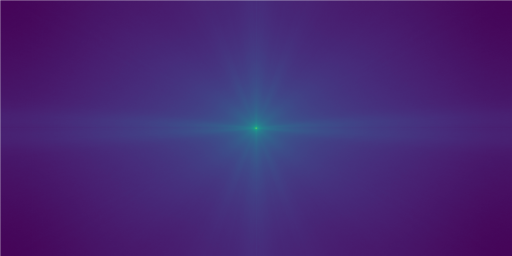} & 
			\includegraphics[width=0.5\columnwidth]{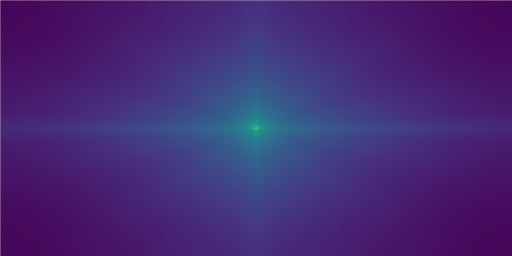}  &  
			\includegraphics[width=0.5\columnwidth]{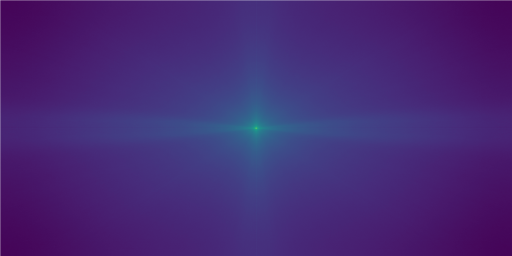} \\
			\textbf{} & \textbf{Rain} & \textbf{} \\
			& 
			\includegraphics[width=0.5\columnwidth]{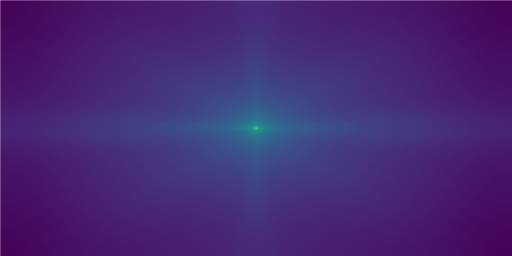}  &  
			\\
		\end{tabular}%
	}
	\caption{Amplitude spectrum in log-scale for Cityscapes the different ACDC subsets and the Cityscapes dataset using Blur and Noise augmentation.}
	\label{fig:fft2}
\end{figure}

In \cref{fig:fft1} and \cref{fig:fft2}, we compare the mean frequency amplitudes of the Cityscapes dataset with the different ACDC subsets.
We observe that ACDC contains much more mid- and high-frequency components in the images, specifically \textit{Snow} and \textit{Rain} contain more higher frequency components. 
From the ACDC subsets \textit{Night} is most similar to \textit{Cityscapes} in the frequency domain, which could explain why forgetting for \textit{Night} is less. 
Furthermore, we also see that blurring and the addition of noise to the image have a significant impact in the frequency domain. 
The goal of adding noise and gaussian blur is to remove the information contained in the high-frequency components of the image so that the model is forced to learn features focusing on low-frequency information that can be reused on the target domain, where the domains are more similar. The plots show that the methods are effectively achieving this. 
However, we observe that learning color-invariant features are much more effective at mitigating forgetting, which we also confirm for other CNN architectures in \cref{res_erf}.

\section{Which BN layers are affected?}
\begin{figure}
	\centering
	\includegraphics[width=.75\columnwidth]{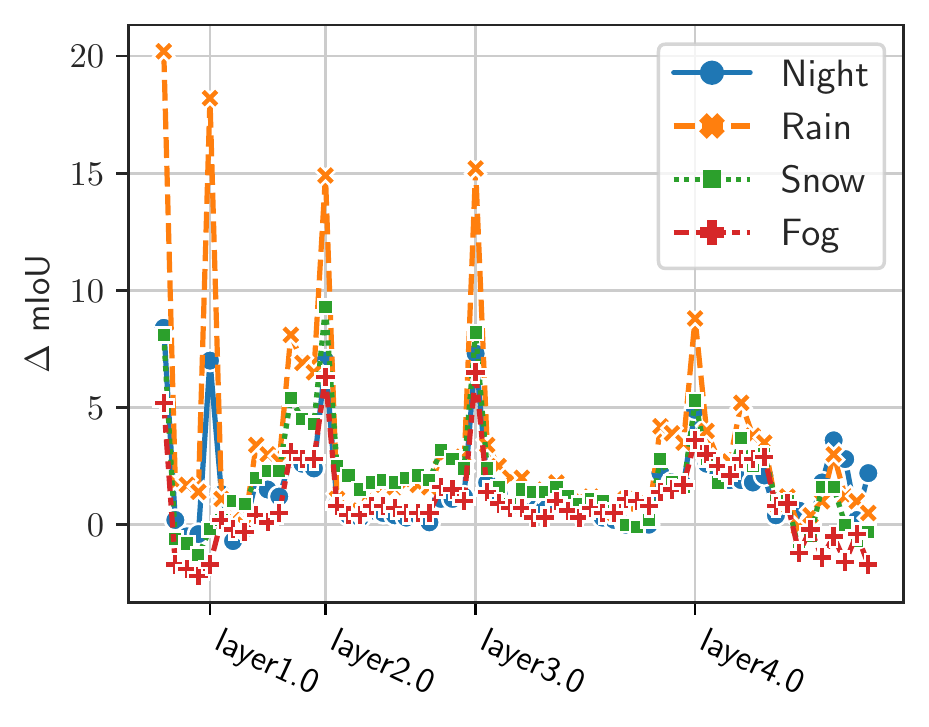}\hfill
	\caption{Change in mIoU on the first task after re-estimation of the population statistics of specific BN layers (horizontal axis). Re-Estimation mostly affects the first BN layer and the last BN layers in each stage's first block.}
	\label{fig:BN_layer_wise}
\end{figure}

In Section 4.1, we found that changing population statistics of BN layers are a significant cause of catastrophic forgetting. 
To study which BN layer is most affected by the changing population statistics, we re-estimate the BN statistics for one layer at a time. 
The results are displayed in \cref{fig:BN_layer_wise}. 
We observe that the first BN layer has the most impact on forgetting and that the last BN layer in the first block of each stage (e.g. \textit{layer2.0.downsample.1}) has a comparable impact when the remaining BN layers are not adjusted. 
These specific layers coincide with blocks that were identified as critical layers by Zhang \etal \cite{alllayerequal}.
Interestingly, in a set of freezing experiments in which we freeze the model up until specific intermediate layers, we observe that the severe activation drift inside the model is always happening in these specific layers. These results are discussed in the \cref{sec:freeze}. \\
Furthermore, we also repeat the re-estimation experiment only for the first BN layer. 
In \cref{tab:bnre_estimate_1} we still observe the same trends as for re-estimating the statistics for all layers, but with slightly reduced improvements. 
These results further demonstrate that the change in population statistics is, in fact, mostly affecting the very first BN layer.

\begin{table}
	\resizebox{\columnwidth}{!}{%
		\begin{tabular}{@{}l|c|cc|ccc}
			\toprule
			\multicolumn{1}{c|}{} & \textbf{CS} & \multicolumn{2}{c|}{\textbf{Rain}} & \multicolumn{3}{c}{\textbf{Night}} \\
			\multicolumn{1}{c|}{} & \textit{Test} & \textit{CS} & \textit{Test} & \textit{CS} & \textit{Rain} & \textit{Test} \\
			\multicolumn{1}{c|}{\multirow{-3}{*}{\textbf{Method}}} & \textit{mIoU} & \textit{forg.} & \textit{mIoU} & \textit{forg.} & \textit{forg.} & \textit{mIoU} \\ \midrule
			FT & \cellcolor[HTML]{FBA476}72.0 & \cellcolor[HTML]{F8696B}33.2 & \cellcolor[HTML]{F8696B}57.7 & \cellcolor[HTML]{F8696B}27.8 & \cellcolor[HTML]{FB9C75}24.9 & \cellcolor[HTML]{FBA877}45.3 \\
			AutoAlb. & \cellcolor[HTML]{FCB379}72.2 & \cellcolor[HTML]{91CB7D}10.7 & \cellcolor[HTML]{FBAE78}59.4 & \cellcolor[HTML]{BCD780}15.2 & \cellcolor[HTML]{8DCA7D}18.2 & \cellcolor[HTML]{F9EA84}47.4 \\
			Distort & \cellcolor[HTML]{FA8E72}71.7 & \cellcolor[HTML]{EBE582}19.0 & \cellcolor[HTML]{FFEB84}60.9 & \cellcolor[HTML]{FED881}20.8 & \cellcolor[HTML]{F8696B}26.6 & \cellcolor[HTML]{F3E884}47.5 \\ \hline
			ImageNet & \cellcolor[HTML]{BED881}73.9 & \cellcolor[HTML]{FFD981}22.5 & \cellcolor[HTML]{FFEB84}60.9 & \cellcolor[HTML]{FA8471}26.1 & \cellcolor[HTML]{FEC97E}23.4 & \cellcolor[HTML]{FCC37C}46.1 \\
			MOCO & \cellcolor[HTML]{63BE7B}75.2 & \cellcolor[HTML]{FCAC78}26.8 & \cellcolor[HTML]{C5DB81}63.5 & \cellcolor[HTML]{EAE582}18.2 & \cellcolor[HTML]{C2D980}20.1 & \cellcolor[HTML]{FEE783}47.2 \\
			DINO & \cellcolor[HTML]{71C37C}75.0 & \cellcolor[HTML]{FED07F}23.4 & \cellcolor[HTML]{B0D580}64.4 & \cellcolor[HTML]{EBE582}18.3 & \cellcolor[HTML]{DEE182}21.1 & \cellcolor[HTML]{6AC07C}49.7 \\ \hline
			CN & \cellcolor[HTML]{F8696B}71.2 & \cellcolor[HTML]{A7D17E}12.7 & \cellcolor[HTML]{F98D72}58.6 & \cellcolor[HTML]{FED380}21.1 & \cellcolor[HTML]{FA7E70}25.9 & \cellcolor[HTML]{F8696B}43.4 \\
			Combined & \cellcolor[HTML]{CCDD82}73.7 & \cellcolor[HTML]{63BE7B}6.4 & \cellcolor[HTML]{63BE7B}67.8 & \cellcolor[HTML]{63BE7B}9.4 & \cellcolor[HTML]{63BE7B}16.7 & \cellcolor[HTML]{63BE7B}49.8 \\ \bottomrule
		\end{tabular}%
	}
	\caption{Results on the sequence of \textit{CS} $\rightarrow$ \textit{Rain} $\rightarrow$ \textit{Night} in mIoU (\%), highlight that pre-training and augmentation can decrease forgetting also in a longer task sequences, reducing forgetting not only for the initial task, but for the intermediate task as well.}
	\label{tab:bnre_estimate_1}
\end{table}

\section{Layer freezing experiments} \label{sec:freeze}
\begin{figure}
	\centering
	\includegraphics[width=\columnwidth]{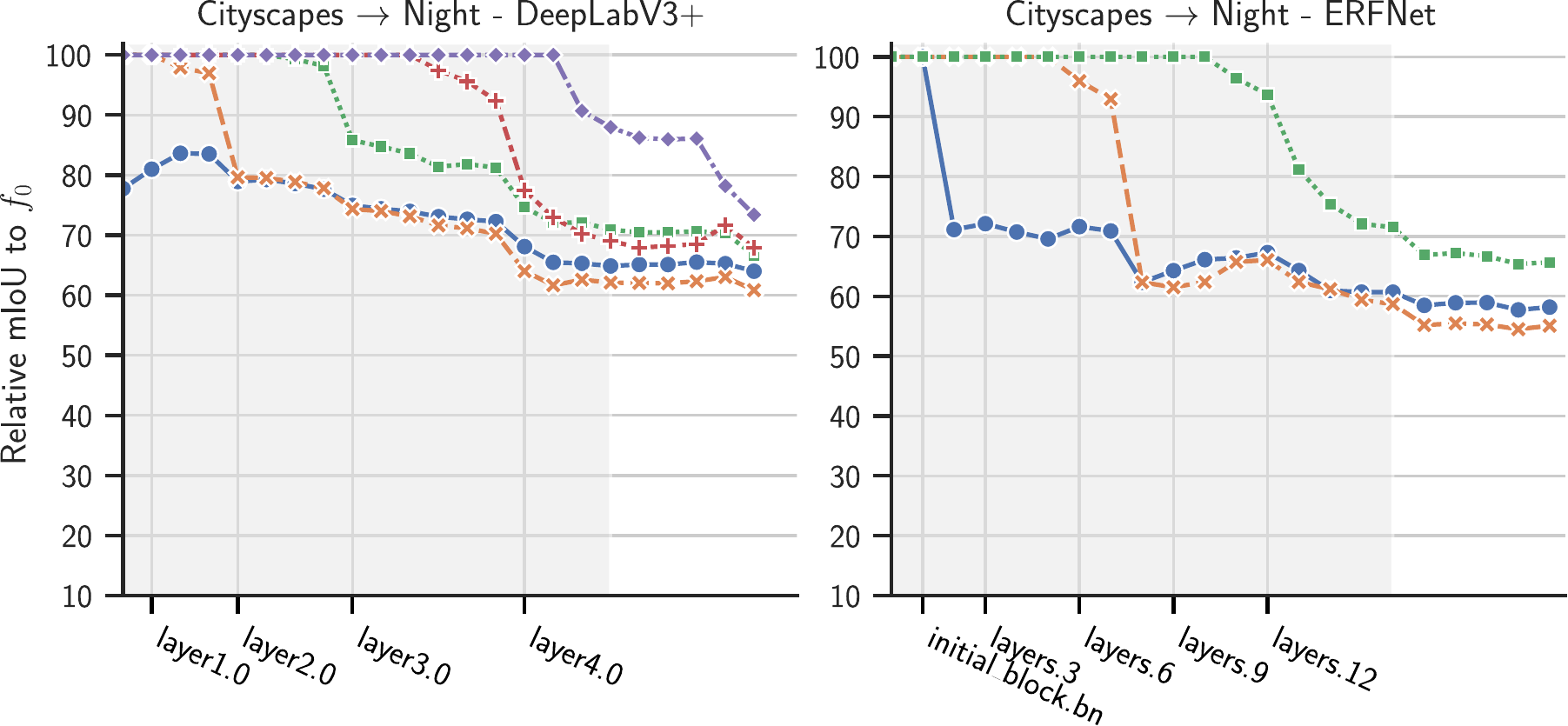}
	\caption{Activation drift between $f_1$ to $f_0$ measured by relative mIoU on the first task of the models stitched together at specific layers (horizontal axis). During training on \textit{Night} we froze layers of ERFNet and DeepLabV3+ starting from the very first block. We see that freezing layers during training on the new task fixes early representation shift, but shifts the initial representation shift to later layers.}
	\label{fig:freeze_erf}
\end{figure}

Previous experiments have shown that a major cause of forgetting is the representation shift in the early layers of the model. So naturally, the question arises: what happens if we just freeze the early layers and fix the population statistics of the BN layers during incremental training? Therefore, in a set of experiments, we freeze an increasing number of layers during training on \textit{Night} and \textit{Rain} subset, starting from the very first layer. The results in \cref{tab:re-train-freeze} show that freezing the first few layers of the encoder has only a minor effect on reducing forgetting or inhibiting learning on the new task. Only when freezing a larger number of layers in the encoder do we observe that the model is less affected by forgetting, but in turn is also inhibited in adapting to \textit{Night}. The reason why the effect is not as prominent for early layers can be seen in the layer stitching plots in \cref{fig:freeze_erf}. The representational shift of the initial layers is shifted to specific later layers, where the similarity drops down to the level of the non-frozen model. The layers where this representation shift occurs coincide with the layers that were most affected by BN re-estimation. 
These results indicate that the low-level feature change cannot be addressed by freezing early layers, as it will inhibit learning or shift the activation drift simply to later layers.

\begin{table}
	\resizebox{\columnwidth}{!}{%
		\begin{tabular}{@{}c|l|ccc|ccc}
			\toprule
			& \multicolumn{1}{c|}{\textbf{}} & \multicolumn{3}{c|}{\textbf{Night}} & \multicolumn{3}{c}{\textbf{Rain}} \\
			& \multicolumn{1}{c|}{\textbf{Frozen}} & \textit{Cityscapes} & \textit{Night} & \textit{} & \textit{Cityscapes} & \textit{Rain} & \textit{} \\
			\multirow{-3}{*}{\textbf{Model}} & \multicolumn{1}{c|}{\textbf{until}} & \textit{mIoU} & \textit{mIoU} & \textit{Forgetting} & \textit{mIoU} & \textit{mIoU} & \textit{Forgetting} \\ \midrule
			&  & \cellcolor[HTML]{FBB179}45.9 & \cellcolor[HTML]{85C87D}43.6 & \cellcolor[HTML]{FCB27A}26.1 & \cellcolor[HTML]{F8716C}38.8 & \cellcolor[HTML]{63BE7B}57.8 & \cellcolor[HTML]{F9726D}33.2 \\
			& layer1.0 & \cellcolor[HTML]{F8696B}43.6 & \cellcolor[HTML]{63BE7B}44.5 & \cellcolor[HTML]{F8696B}28.4 & \cellcolor[HTML]{F8696B}38.5 & \cellcolor[HTML]{81C77D}57.5 & \cellcolor[HTML]{F8696B}33.5 \\
			& layer2.1 & \cellcolor[HTML]{FFEB84}47.7 & \cellcolor[HTML]{A7D27F}42.7 & \cellcolor[HTML]{FFEB84}24.3 & \cellcolor[HTML]{FFEB84}42.9 & \cellcolor[HTML]{FFEB84}56.2 & \cellcolor[HTML]{FFEB84}29.1 \\
			& layer3.2 & \cellcolor[HTML]{DDE283}48.9 & \cellcolor[HTML]{FDD47F}39.0 & \cellcolor[HTML]{DCE182}23.1 & \cellcolor[HTML]{B7D780}50.4 & \cellcolor[HTML]{FCB87A}52.6 & \cellcolor[HTML]{B6D67F}21.6 \\
			\multirow{-5}{*}{\rotatebox[origin=c]{90}{\parbox[c]{1cm}{\centering Deep-LabV3+.}}} & layer4.1 & \cellcolor[HTML]{63BE7B}53.1 & \cellcolor[HTML]{F8696B}32.5 & \cellcolor[HTML]{63BE7B}18.9 & \cellcolor[HTML]{63BE7B}59.1 & \cellcolor[HTML]{F8696B}47.0 & \cellcolor[HTML]{63BE7B}12.9 \\ \hline
			&  & \cellcolor[HTML]{F98A71}37.1 & \cellcolor[HTML]{63BE7B}41.7 & \cellcolor[HTML]{FA8B72}31.3 & \cellcolor[HTML]{F2E884}31.9 & \cellcolor[HTML]{FFEB84}53.7 & \cellcolor[HTML]{F1E783}36.5 \\
			& initial.bn & \cellcolor[HTML]{ECE683}38.8 & \cellcolor[HTML]{80C77D}41.3 & \cellcolor[HTML]{EBE582}29.6 & \cellcolor[HTML]{FCB479}28.6 & \cellcolor[HTML]{DDE283}54.6 & \cellcolor[HTML]{FDB57A}39.8 \\
			& layers.5 & \cellcolor[HTML]{F8696B}36.8 & \cellcolor[HTML]{FA9A74}37.7 & \cellcolor[HTML]{F8696B}31.6 & \cellcolor[HTML]{F8696B}26.3 & \cellcolor[HTML]{FEDB81}52.9 & \cellcolor[HTML]{F8696B}42.1 \\
			\multirow{-4}{*}{\rotatebox[origin=c]{90}{\parbox[c]{1cm}{\centering ERFNet}}} & layers.10 & \cellcolor[HTML]{63BE7B}44.9 & \cellcolor[HTML]{F8696B}36.6 & \cellcolor[HTML]{63BE7B}23.5 & \cellcolor[HTML]{63BE7B}49.6 & \cellcolor[HTML]{F98B71}48.8 & \cellcolor[HTML]{63BE7B}18.8 \\ \bottomrule
		\end{tabular}%
	}
	\caption{Performance and forgetting in mIoU [\%] on \textit{CS} and the ACDC subsets \textit{Night} and \textit{Rain}, when the early layers of the models are frozen during fine-tuning to the new task. }
	\label{tab:re-train-freeze}
	
\end{table}

\section{Longer task sequence}
We evaluate the training schemes also on a multi-step domain increment with \textit{CS}, \textit{Rain} and \textit{Night}, where augmentations are again only used during training on CS. \cref{tab:seq} shows that pre-training and augmentation can decrease forgetting also in a longer task sequences, reducing forgetting not only for the initial task, but for the intermediate task as well. 
This indicates that once general low-level features are learned, their benefits remain even after the model is fine-tuned on a new domain without the additional augmentations.
However, we note that the interaction between these domains can be intricate, as we observe a reduction in forgetting on CS after the model was trained on \textit{Night} when no augmentations are used. 
Furthermore, we also noticed in our preliminary experiments that the order or similarity of the tasks further impacts the severity of forgetting. 

\begin{table}
	\resizebox{\columnwidth}{!}{%
		\begin{tabular}{@{}l|c|cc|ccc}
			\toprule
			\multicolumn{1}{c|}{} & \textbf{CS} & \multicolumn{2}{c|}{\textbf{Rain}} & \multicolumn{3}{c}{\textbf{Night}} \\
			\multicolumn{1}{c|}{} & \textit{Test} & \multicolumn{1}{c|}{\textit{CS}} & \textit{Test} & \multicolumn{1}{c|}{\textit{CS}} & \multicolumn{1}{c|}{\textit{Rain}} & \textit{Test} \\
			\multicolumn{1}{c|}{\multirow{-3}{*}{\textbf{Method}}} & \textit{mIoU} & \multicolumn{1}{c|}{\textit{forg.}} & \textit{mIoU} & \multicolumn{1}{c|}{\textit{forg.}} & \multicolumn{1}{c|}{\textit{forg.}} & \textit{mIoU} \\ \midrule
			FT & \cellcolor[HTML]{FBA476}72.0 & \multicolumn{1}{c|}{\cellcolor[HTML]{F8696B}33.2} & \cellcolor[HTML]{F8696B}57.7 & \multicolumn{1}{c|}{\cellcolor[HTML]{F8696B}27.8} & \multicolumn{1}{c|}{\cellcolor[HTML]{FB9C75}24.9} & \cellcolor[HTML]{FBA877}45.3 \\
			AutoAlb. & \cellcolor[HTML]{FCB379}72.2 & \multicolumn{1}{c|}{\cellcolor[HTML]{91CB7D}10.7} & \cellcolor[HTML]{FBAE78}59.4 & \multicolumn{1}{c|}{\cellcolor[HTML]{BCD780}15.2} & \multicolumn{1}{c|}{\cellcolor[HTML]{8DCA7D}18.2} & \cellcolor[HTML]{F9EA84}47.4 \\
			Distort & \cellcolor[HTML]{FA8E72}71.7 & \multicolumn{1}{c|}{\cellcolor[HTML]{EBE582}19.0} & \cellcolor[HTML]{FFEB84}60.9 & \multicolumn{1}{c|}{\cellcolor[HTML]{FED881}20.8} & \multicolumn{1}{c|}{\cellcolor[HTML]{F8696B}26.6} & \cellcolor[HTML]{F3E884}47.5 \\ \hline
			ImageNet & \cellcolor[HTML]{BED881}73.9 & \multicolumn{1}{c|}{\cellcolor[HTML]{FFD981}22.5} & \cellcolor[HTML]{FFEB84}60.9 & \multicolumn{1}{c|}{\cellcolor[HTML]{FA8471}26.1} & \multicolumn{1}{c|}{\cellcolor[HTML]{FEC97E}23.4} & \cellcolor[HTML]{FCC37C}46.1 \\
			MOCO & \cellcolor[HTML]{63BE7B}75.2 & \multicolumn{1}{c|}{\cellcolor[HTML]{FCAC78}26.8} & \cellcolor[HTML]{C5DB81}63.5 & \multicolumn{1}{c|}{\cellcolor[HTML]{EAE582}18.2} & \multicolumn{1}{c|}{\cellcolor[HTML]{C2D980}20.1} & \cellcolor[HTML]{FEE783}47.2 \\
			DINO & \cellcolor[HTML]{71C37C}75.0 & \multicolumn{1}{c|}{\cellcolor[HTML]{FED07F}23.4} & \cellcolor[HTML]{B0D580}64.4 & \multicolumn{1}{c|}{\cellcolor[HTML]{EBE582}18.3} & \multicolumn{1}{c|}{\cellcolor[HTML]{DEE182}21.1} & \cellcolor[HTML]{6AC07C}49.7 \\ \hline
			CN & \cellcolor[HTML]{F8696B}71.2 & \multicolumn{1}{c|}{\cellcolor[HTML]{A7D17E}12.7} & \cellcolor[HTML]{F98D72}58.6 & \multicolumn{1}{c|}{\cellcolor[HTML]{FED380}21.1} & \multicolumn{1}{c|}{\cellcolor[HTML]{FA7E70}25.9} & \cellcolor[HTML]{F8696B}43.4 \\
			Combined & \cellcolor[HTML]{CCDD82}73.7 & \multicolumn{1}{c|}{\cellcolor[HTML]{63BE7B}6.4} & \cellcolor[HTML]{63BE7B}67.8 & \multicolumn{1}{c|}{\cellcolor[HTML]{63BE7B}9.4} & \multicolumn{1}{c|}{\cellcolor[HTML]{63BE7B}16.7} & \cellcolor[HTML]{63BE7B}49.8 \\ \bottomrule
		\end{tabular}%
	}
	\caption{Results for CS $\rightarrow$ \textit{Rain}$\rightarrow$ \textit{Night} with DeepLabV3+. We see that the combination of pre-training with DINO, AutoAlbum and Continual Normalization (denoted as \textit{Combined}) drastically decreases forgetting even in longer task sequences.}
	\label{tab:seq}
\end{table}

\section{Ablation on architectures} \label{res_erf}
We validate our results on the effect of pre-training and augmentation observed previously on DeepLabv3+ also for ERFNet~\cite{erfnet}, BiSeNet V2~\cite{yu2021bisenet}, HRNetV2~\cite{WangSCJDZLMTWLX19} and RTFormer~\cite{wang2022rtformer} in \cref{tab:pre-train_erfnet,tab:pre-train_bise,tab:pre-train_hrnet}. We select these networks as they have very distinct architectures compared to DeepLabV3+. HRNetV2 and BiSeNet V2 use multiple parallel branches, ERFNet has a significantly lower number of parameters, and RTFormer is a computationally efficient transformer-based model.
\cref{tab:pre-train_erfnet,tab:pre-train_bise,tab:pre-train_hrnet} show that augmentations and pre-training also significantly reduce forgetting for those selected architectures. Specifically, we see that the combination of pre-training and AutoAlbum leads to significant improvements for all models across all datasets.
Furthermore, we see that ERFNet and BiSeNet V2 are much more affected by catastrophic forgetting due to their much smaller size. 
However, besides this difference, we overall see very similar results, as Distortion and AutoAlbum are the most effective methods to enforce effective feature reuse and thus a reduction of forgetting. 
Moreover, we make the same observations for ImageNet pre-training, where we achieve higher mIoU on the target dataset but are not as effective at reducing forgetting compared to the models trained with augmentation. 
The only noticeable difference between the results of BiSeNet V2, ERFNet and DeepLabv3+ is the worse performance on \textit{Snow}, which is drastically worse than the performance of the different subsets, although we use the same training regime as before.
Finally, for RTFormer-Base we surprisingly discover results that are similar to CNN architectures than to the results of SegFormer. 
We hypothesize that this is caused by the use of Batch Normalization instead of layer normalization in the Encoder of RTFormer.
These results, combined with the observation that SegFormer is less affected by the domain shift, demonstrate that while our results are applicable to different CNN architectures using BN, catastrophic forgetting significantly varies between architectures, as previous work has pointed out \cite{archmatters, kalb2023effects}.
\begin{table*}
	\resizebox{\textwidth}{!}{%
		\begin{tabular}{@{}lccccccccccccc}
			\multicolumn{14}{c}{\textbf{ERFNet}} \\ \midrule
			\multicolumn{1}{c|}{} & \multicolumn{1}{c|}{\textbf{Cityscapes}} & \multicolumn{3}{c|}{\textbf{Night}} & \multicolumn{3}{c|}{\textbf{Rain}} & \multicolumn{3}{c|}{\textbf{Fog}} & \multicolumn{3}{c}{\textbf{Snow}} \\
			\multicolumn{1}{c|}{} & \multicolumn{1}{c|}{\textit{Test}} & \textit{Zero} & \textit{Test} & \multicolumn{1}{c|}{\textit{}} & \textit{Zero} & \multicolumn{1}{c|}{\textit{Test}} & \multicolumn{1}{c|}{\textit{}} & \textit{Zero} & \textit{Test} & \multicolumn{1}{c|}{\textit{}} & \textit{Zero} & \textit{Test} & \textit{} \\
			\multicolumn{1}{c|}{\multirow{-3}{*}{\textbf{Method}}} & \multicolumn{1}{c|}{\textit{mIoU}} & \textit{Shot} & \textit{mIoU} & \multicolumn{1}{c|}{\textit{Forgetting}} & \textit{Shot} & \multicolumn{1}{c|}{\textit{mIoU}} & \multicolumn{1}{c|}{\textit{Forgetting}} & \textit{Shot} & \textit{mIoU} & \multicolumn{1}{c|}{\textit{Forgetting}} & \textit{Shot} & \textit{mIoU} & \textit{Forgetting} \\ \midrule
			\multicolumn{1}{l|}{FT} & \multicolumn{1}{c|}{\cellcolor[HTML]{FED980}68.4} & \cellcolor[HTML]{FDCC7E}8.2 & \cellcolor[HTML]{FCC47C}41.7 & \multicolumn{1}{c|}{\cellcolor[HTML]{F8696B}31.3} & \cellcolor[HTML]{FA9172}19.5 & \multicolumn{1}{c|}{\cellcolor[HTML]{FDD680}53.7} & \multicolumn{1}{c|}{\cellcolor[HTML]{FFDF82}36.5} & \cellcolor[HTML]{F97C6E}15.2 & \cellcolor[HTML]{FDCD7E}58.0 & \multicolumn{1}{c|}{\cellcolor[HTML]{F8696B}35.3} & \cellcolor[HTML]{F98670}9.8 & \cellcolor[HTML]{FCB479}57.1 & \cellcolor[HTML]{F9766E}57.4 \\
			\multicolumn{1}{l|}{AutoAlb.} & \multicolumn{1}{c|}{\cellcolor[HTML]{F8696B}64.0} & \cellcolor[HTML]{A3D17F}14.4 & \cellcolor[HTML]{FEE783}42.6 & \multicolumn{1}{c|}{\cellcolor[HTML]{CDDC81}18.9} & \cellcolor[HTML]{B7D780}30.5 & \multicolumn{1}{c|}{\cellcolor[HTML]{EAE583}54.4} & \multicolumn{1}{c|}{\cellcolor[HTML]{8AC97D}14.7} & \cellcolor[HTML]{B1D580}32.9 & \cellcolor[HTML]{F98A71}56.4 & \multicolumn{1}{c|}{\cellcolor[HTML]{A3D07E}16.4} & \cellcolor[HTML]{BFD981}22.7 & \cellcolor[HTML]{F8696B}55.7 & \cellcolor[HTML]{BCD780}25.1 \\
			\multicolumn{1}{l|}{Distort} & \multicolumn{1}{c|}{\cellcolor[HTML]{FA9473}65.7} & \cellcolor[HTML]{63BE7B}17.7 & \cellcolor[HTML]{FFEB84}42.7 & \multicolumn{1}{c|}{\cellcolor[HTML]{D1DD81}19.3} & \cellcolor[HTML]{B1D580}31.0 & \multicolumn{1}{c|}{\cellcolor[HTML]{FA9172}52.5} & \multicolumn{1}{c|}{\cellcolor[HTML]{9CCE7E}18.0} & \cellcolor[HTML]{A3D17F}34.9 & \cellcolor[HTML]{FEE282}58.5 & \multicolumn{1}{c|}{\cellcolor[HTML]{BAD780}19.4} & \cellcolor[HTML]{A7D27F}25.3 & \cellcolor[HTML]{F8696B}55.7 & \cellcolor[HTML]{B0D47F}22.6 \\
			\multicolumn{1}{l|}{Gaus} & \multicolumn{1}{c|}{\cellcolor[HTML]{F9826F}65.0} & \cellcolor[HTML]{FA9F75}6.1 & \cellcolor[HTML]{FA9373}40.4 & \multicolumn{1}{c|}{\cellcolor[HTML]{FCB179}27.3} & \cellcolor[HTML]{F8696B}17.3 & \multicolumn{1}{c|}{\cellcolor[HTML]{F6E984}54.2} & \multicolumn{1}{c|}{\cellcolor[HTML]{FCA176}41.4} & \cellcolor[HTML]{F8696B}14.1 & \cellcolor[HTML]{FCC57C}57.8 & \multicolumn{1}{c|}{\cellcolor[HTML]{FFEB84}28.4} & \cellcolor[HTML]{F8696B}8.1 & \cellcolor[HTML]{F8796E}56.0 & \cellcolor[HTML]{FECE7F}43.2 \\
			\multicolumn{1}{l|}{Noise} & \multicolumn{1}{c|}{\cellcolor[HTML]{F98C71}65.4} & \cellcolor[HTML]{F8696B}3.6 & \cellcolor[HTML]{FFEB84}42.7 & \multicolumn{1}{c|}{\cellcolor[HTML]{FCA877}27.8} & \cellcolor[HTML]{FBAA77}20.8 & \multicolumn{1}{c|}{\cellcolor[HTML]{F8696B}51.8} & \multicolumn{1}{c|}{\cellcolor[HTML]{FED07F}37.7} & \cellcolor[HTML]{FCBA7A}18.6 & \cellcolor[HTML]{F8696B}55.6 & \multicolumn{1}{c|}{\cellcolor[HTML]{FB9774}32.9} & \cellcolor[HTML]{FFEB84}15.6 & \cellcolor[HTML]{FA8E72}56.4 & \cellcolor[HTML]{FCA577}49.8 \\ \hline
			\multicolumn{1}{l|}{ImageNet} & \multicolumn{1}{c|}{\cellcolor[HTML]{B4D680}70.4} & \cellcolor[HTML]{EAE583}10.7 & \cellcolor[HTML]{FEEB84}42.8 & \multicolumn{1}{c|}{\cellcolor[HTML]{FB9273}29.0} & \cellcolor[HTML]{EFE784}25.7 & \multicolumn{1}{c|}{\cellcolor[HTML]{82C77D}56.1} & \multicolumn{1}{c|}{\cellcolor[HTML]{FFE383}36.2} & \cellcolor[HTML]{DFE283}26.1 & \cellcolor[HTML]{78C47D}64.6 & \multicolumn{1}{c|}{\cellcolor[HTML]{FECB7E}30.1} & \cellcolor[HTML]{EBE683}17.8 & \cellcolor[HTML]{DEE283}58.6 & \cellcolor[HTML]{F8696B}59.5 \\
			\multicolumn{1}{l|}{MOCO} & \multicolumn{1}{c|}{\cellcolor[HTML]{63BE7B}71.8} & \cellcolor[HTML]{F4E884}10.2 & \cellcolor[HTML]{FCEB84}43.0 & \multicolumn{1}{c|}{\cellcolor[HTML]{FB9D75}28.4} & \cellcolor[HTML]{FCBA7A}21.7 & \multicolumn{1}{c|}{\cellcolor[HTML]{94CD7E}55.8} & \multicolumn{1}{c|}{\cellcolor[HTML]{FBE983}34.9} & \cellcolor[HTML]{FFEB84}21.3 & \cellcolor[HTML]{BBD881}61.7 & \multicolumn{1}{c|}{\cellcolor[HTML]{FDC67D}30.4} & \cellcolor[HTML]{FDCF7E}14.0 & \cellcolor[HTML]{63BE7B}60.4 & \cellcolor[HTML]{FFEB84}38.4 \\
			\multicolumn{1}{l|}{DINO} & \multicolumn{1}{c|}{\cellcolor[HTML]{C6DB81}70.1} & \cellcolor[HTML]{FCBF7B}7.6 & \cellcolor[HTML]{F9EA84}43.3 & \multicolumn{1}{c|}{\cellcolor[HTML]{FDC37D}26.3} & \cellcolor[HTML]{FFEB84}24.3 & \multicolumn{1}{c|}{\cellcolor[HTML]{63BE7B}56.6} & \multicolumn{1}{c|}{\cellcolor[HTML]{F8696B}45.8} & \cellcolor[HTML]{FEE182}20.8 & \cellcolor[HTML]{FBEA84}58.9 & \multicolumn{1}{c|}{\cellcolor[HTML]{FDC07C}30.7} & \cellcolor[HTML]{FFEB84}15.6 & \cellcolor[HTML]{9ACE7F}59.6 & \cellcolor[HTML]{FDB77A}46.9 \\
			\multicolumn{1}{l|}{CN} & \multicolumn{1}{c|}{\cellcolor[HTML]{B4D680}70.4} & \cellcolor[HTML]{FFEB84}9.6 & \cellcolor[HTML]{FA9373}40.4 & \multicolumn{1}{c|}{\cellcolor[HTML]{E8E482}21.7} & \cellcolor[HTML]{DAE182}27.5 & \multicolumn{1}{c|}{\cellcolor[HTML]{FA9D75}52.7} & \multicolumn{1}{c|}{\cellcolor[HTML]{8ECA7D}15.4} & \cellcolor[HTML]{D3DF82}27.8 & \cellcolor[HTML]{B6D680}61.9 & \multicolumn{1}{c|}{\cellcolor[HTML]{AED37F}17.9} & \cellcolor[HTML]{FBB078}12.2 & \cellcolor[HTML]{A1D07F}59.5 & \cellcolor[HTML]{A7D17E}20.8 \\
			\multicolumn{1}{l|}{Combined} & \multicolumn{1}{c|}{\cellcolor[HTML]{D7E082}69.8} & \cellcolor[HTML]{D9E082}11.6 & \cellcolor[HTML]{FAEA84}43.2 & \multicolumn{1}{c|}{\cellcolor[HTML]{A7D17E}15.0} & \cellcolor[HTML]{63BE7B}37.6 & \multicolumn{1}{c|}{\cellcolor[HTML]{63BE7B}57.5} & \multicolumn{1}{c|}{\cellcolor[HTML]{63BE7B}8.0} & \cellcolor[HTML]{63BE7B}44.3 & \cellcolor[HTML]{63BE7B}65.5 & \multicolumn{1}{c|}{\cellcolor[HTML]{7CC57C}11.3} & \cellcolor[HTML]{63BE7B}32.7 & \cellcolor[HTML]{8CCA7E}59.8 & \cellcolor[HTML]{95CC7D}17.2 \\
			\multicolumn{1}{l|}{Replay} & \multicolumn{1}{c|}{\cellcolor[HTML]{63BE7B}68.4} & \cellcolor[HTML]{FDCC7E}8.2 & \cellcolor[HTML]{F8696B}39.3 & \multicolumn{1}{c|}{\cellcolor[HTML]{6AC07B}8.8} & \cellcolor[HTML]{FA9172}19.5 & \multicolumn{1}{c|}{\cellcolor[HTML]{FEE282}53.9} & \multicolumn{1}{c|}{\cellcolor[HTML]{63BE7B}7.7} & \cellcolor[HTML]{F97C6E}15.2 & \cellcolor[HTML]{FFEB84}58.7 & \multicolumn{1}{c|}{\cellcolor[HTML]{63BE7B}8.0} & \cellcolor[HTML]{F98670}9.8 & \cellcolor[HTML]{FFEB84}58.1 & \cellcolor[HTML]{63BE7B}7.2 \\
			\multicolumn{1}{l|}{Offline} & \multicolumn{1}{c|}{} & 40.1 & 43.1 & \multicolumn{1}{c|}{15.6} & 50.5 & \multicolumn{1}{c|}{55.1} & \multicolumn{1}{c|}{19.9} & 58.1 & 61.5 & \multicolumn{1}{c|}{14.9} & 53.6 & 55.8 & 23.3 \\ \bottomrule
		\end{tabular}%
	}
	\caption{Results of ERFNet \cite{erfnet} on \textit{CS} $\rightarrow$ \textit{ACDC} in mIoU (\%) for each subset of ACDC using different pre-training and augmentations strategies (Augment.). Compared to DeepLabV3+, ERFNet is much more affected by Forgetting, specifically on \textit{Snow}. However, Augmentations and pretraining show the same effects as for the experiments in the main paper.}
	\label{tab:pre-train_erfnet}
\end{table*}

\begin{table*}
	\resizebox{\textwidth}{!}{%
		\begin{tabular}{@{}lccccccccccccc}
			
			\multicolumn{14}{c}{\textbf{BiSeNet V2}} \\ \midrule
			\multicolumn{1}{c|}{} & \multicolumn{1}{c|}{\textbf{Cityscapes}} & \multicolumn{3}{c|}{\textbf{Night}} & \multicolumn{3}{c|}{\textbf{Rain}} & \multicolumn{3}{c|}{\textbf{Fog}} & \multicolumn{3}{c}{\textbf{Snow}} \\
			\multicolumn{1}{c|}{} & \multicolumn{1}{c|}{\textit{Test}} & \multicolumn{1}{c|}{\textit{Zero}} & \multicolumn{1}{c|}{\textit{Test}} & \multicolumn{1}{c|}{\textit{}} & \multicolumn{1}{c|}{\textit{Zero}} & \multicolumn{1}{c|}{\textit{Test}} & \multicolumn{1}{c|}{\textit{}} & \multicolumn{1}{c|}{\textit{Zero}} & \multicolumn{1}{c|}{\textit{Test}} & \multicolumn{1}{c|}{\textit{}} & \multicolumn{1}{c|}{\textit{Zero}} & \multicolumn{1}{c|}{\textit{Test}} & \textit{} \\
			\multicolumn{1}{c|}{\multirow{-3}{*}{\textbf{Method}}} & \multicolumn{1}{c|}{\textit{mIoU}} & \multicolumn{1}{c|}{\textit{Shot}} & \multicolumn{1}{c|}{\textit{mIoU}} & \multicolumn{1}{c|}{\textit{Forgetting}} & \multicolumn{1}{c|}{\textit{Shot}} & \multicolumn{1}{c|}{\textit{mIoU}} & \multicolumn{1}{c|}{\textit{Forgetting}} & \multicolumn{1}{c|}{\textit{Shot}} & \multicolumn{1}{c|}{\textit{mIoU}} & \multicolumn{1}{c|}{\textit{Forgetting}} & \multicolumn{1}{c|}{\textit{Shot}} & \multicolumn{1}{c|}{\textit{mIoU}} & \textit{Forgetting} \\ \midrule
			\multicolumn{1}{l|}{FT} & \multicolumn{1}{c|}{\cellcolor[HTML]{FDCE7E}67.5} & \multicolumn{1}{c|}{\cellcolor[HTML]{FBA476}4.9} & \multicolumn{1}{c|}{\cellcolor[HTML]{E0E283}41.2} & \multicolumn{1}{c|}{\cellcolor[HTML]{FA8D72}33.7} & \multicolumn{1}{c|}{\cellcolor[HTML]{F97D6E}18.8} & \multicolumn{1}{c|}{\cellcolor[HTML]{A5D17E}52.1} & \multicolumn{1}{c|}{\cellcolor[HTML]{FCA477}40.7} & \multicolumn{1}{c|}{\cellcolor[HTML]{F8796E}14.7} & \multicolumn{1}{c|}{\cellcolor[HTML]{FDD780}57.3} & \multicolumn{1}{c|}{\cellcolor[HTML]{FCAE79}39.4} & \multicolumn{1}{c|}{\cellcolor[HTML]{F8696B}9.3} & \multicolumn{1}{c|}{\cellcolor[HTML]{F1E784}58.1} & \cellcolor[HTML]{FA8170}58.9 \\
			\multicolumn{1}{l|}{AutoAlb.} & \multicolumn{1}{c|}{\cellcolor[HTML]{F8696B}66.6} & \multicolumn{1}{c|}{\cellcolor[HTML]{88C97E}12.8} & \multicolumn{1}{c|}{\cellcolor[HTML]{F5E884}41.0} & \multicolumn{1}{c|}{\cellcolor[HTML]{ECE582}26.2} & \multicolumn{1}{c|}{\cellcolor[HTML]{74C37C}35.5} & \multicolumn{1}{c|}{\cellcolor[HTML]{FDC07C}53.5} & \multicolumn{1}{c|}{\cellcolor[HTML]{D7DF81}23.5} & \multicolumn{1}{c|}{\cellcolor[HTML]{82C77D}39.3} & \multicolumn{1}{c|}{\cellcolor[HTML]{8BCA7E}60.2} & \multicolumn{1}{c|}{\cellcolor[HTML]{FFD981}33.8} & \multicolumn{1}{c|}{\cellcolor[HTML]{81C77D}27.1} & \multicolumn{1}{c|}{\cellcolor[HTML]{FCC57C}56.6} & \cellcolor[HTML]{FECD7F}46.1 \\
			\multicolumn{1}{l|}{Distort} & \multicolumn{1}{c|}{\cellcolor[HTML]{D7E082}68.2} & \multicolumn{1}{c|}{\cellcolor[HTML]{63BE7B}14.8} & \multicolumn{1}{c|}{\cellcolor[HTML]{63BE7B}42.4} & \multicolumn{1}{c|}{\cellcolor[HTML]{FED380}29.7} & \multicolumn{1}{c|}{\cellcolor[HTML]{98CE7F}32.9} & \multicolumn{1}{c|}{\cellcolor[HTML]{FFEB84}52.9} & \multicolumn{1}{c|}{\cellcolor[HTML]{FDC67D}35.3} & \multicolumn{1}{c|}{\cellcolor[HTML]{8BCA7E}38.0} & \multicolumn{1}{c|}{\cellcolor[HTML]{FEE582}58.1} & \multicolumn{1}{c|}{\cellcolor[HTML]{EFE683}29.2} & \multicolumn{1}{c|}{\cellcolor[HTML]{B0D580}23.0} & \multicolumn{1}{c|}{\cellcolor[HTML]{D5DF82}58.3} & \cellcolor[HTML]{E6E382}35.8 \\
			\multicolumn{1}{l|}{Gaus} & \multicolumn{1}{c|}{\cellcolor[HTML]{FBA175}67.1} & \multicolumn{1}{c|}{\cellcolor[HTML]{F8696B}3.8} & \multicolumn{1}{c|}{\cellcolor[HTML]{FEE783}40.8} & \multicolumn{1}{c|}{\cellcolor[HTML]{FA8471}34.2} & \multicolumn{1}{c|}{\cellcolor[HTML]{F8696B}17.6} & \multicolumn{1}{c|}{\cellcolor[HTML]{FFEB84}52.9} & \multicolumn{1}{c|}{\cellcolor[HTML]{FBA176}41.2} & \multicolumn{1}{c|}{\cellcolor[HTML]{F8696B}13.9} & \multicolumn{1}{c|}{\cellcolor[HTML]{C0D981}59.4} & \multicolumn{1}{c|}{\cellcolor[HTML]{F8696B}48.1} & \multicolumn{1}{c|}{\cellcolor[HTML]{F98971}11.0} & \multicolumn{1}{c|}{\cellcolor[HTML]{63BE7B}59.1} & \cellcolor[HTML]{FA8170}58.9 \\
			\multicolumn{1}{l|}{ImageNet} & \multicolumn{1}{c|}{\cellcolor[HTML]{63BE7B}69.5} & \multicolumn{1}{c|}{\cellcolor[HTML]{F1E784}7.0} & \multicolumn{1}{c|}{\cellcolor[HTML]{83C77D}42.1} & \multicolumn{1}{c|}{\cellcolor[HTML]{F8696B}35.7} & \multicolumn{1}{c|}{\cellcolor[HTML]{FA9473}20.2} & \multicolumn{1}{c|}{\cellcolor[HTML]{F8696B}54.7} & \multicolumn{1}{c|}{\cellcolor[HTML]{F8696B}49.9} & \multicolumn{1}{c|}{\cellcolor[HTML]{F86B6B}14.0} & \multicolumn{1}{c|}{\cellcolor[HTML]{63BE7B}60.8} & \multicolumn{1}{c|}{\cellcolor[HTML]{F9786E}46.3} & \multicolumn{1}{c|}{\cellcolor[HTML]{FCBB7A}13.7} & \multicolumn{1}{c|}{\cellcolor[HTML]{FEE883}57.9} & \cellcolor[HTML]{F8696B}62.8 \\ \hline
			\multicolumn{1}{l|}{CN} & \multicolumn{1}{c|}{\cellcolor[HTML]{ABD380}68.7} & \multicolumn{1}{c|}{\cellcolor[HTML]{FCBF7B}5.4} & \multicolumn{1}{c|}{\cellcolor[HTML]{F8696B}37.0} & \multicolumn{1}{c|}{\cellcolor[HTML]{F2E783}26.9} & \multicolumn{1}{c|}{\cellcolor[HTML]{BAD780}30.4} & \multicolumn{1}{c|}{\cellcolor[HTML]{63BE7B}51.5} & \multicolumn{1}{c|}{\cellcolor[HTML]{B2D47F}18.0} & \multicolumn{1}{c|}{\cellcolor[HTML]{DCE182}25.5} & \multicolumn{1}{c|}{\cellcolor[HTML]{FBAD78}54.8} & \multicolumn{1}{c|}{\cellcolor[HTML]{C6DA80}23.1} & \multicolumn{1}{c|}{\cellcolor[HTML]{E2E383}18.7} & \multicolumn{1}{c|}{\cellcolor[HTML]{F98971}54.4} & \cellcolor[HTML]{B4D57F}25.4 \\
			\multicolumn{1}{l|}{Combined} & \multicolumn{1}{c|}{\cellcolor[HTML]{E9E583}68.0} & \multicolumn{1}{c|}{\cellcolor[HTML]{79C57D}13.6} & \multicolumn{1}{c|}{\cellcolor[HTML]{FA9E75}38.6} & \multicolumn{1}{c|}{\cellcolor[HTML]{C4DA80}21.7} & \multicolumn{1}{c|}{\cellcolor[HTML]{63BE7B}36.7} & \multicolumn{1}{c|}{\cellcolor[HTML]{FED680}53.2} & \multicolumn{1}{c|}{\cellcolor[HTML]{95CC7D}13.7} & \multicolumn{1}{c|}{\cellcolor[HTML]{63BE7B}44.0} & \multicolumn{1}{c|}{\cellcolor[HTML]{E8E583}58.8} & \multicolumn{1}{c|}{\cellcolor[HTML]{A0CF7E}17.5} & \multicolumn{1}{c|}{\cellcolor[HTML]{63BE7B}29.6} & \multicolumn{1}{c|}{\cellcolor[HTML]{F8696B}53.2} & \cellcolor[HTML]{A5D17E}22.2 \\
			\multicolumn{1}{l|}{Replay} & \multicolumn{1}{c|}{\cellcolor[HTML]{FDCE7E}67.5} & \multicolumn{1}{c|}{\cellcolor[HTML]{FBA476}4.9} & \multicolumn{1}{c|}{\cellcolor[HTML]{FDCD7E}40.0} & \multicolumn{1}{c|}{\cellcolor[HTML]{63BE7B}10.7} & \multicolumn{1}{c|}{\cellcolor[HTML]{F97D6E}18.8} & \multicolumn{1}{c|}{\cellcolor[HTML]{6EC17B}51.6} & \multicolumn{1}{c|}{\cellcolor[HTML]{63BE7B}6.2} & \multicolumn{1}{c|}{\cellcolor[HTML]{F8796E}14.7} & \multicolumn{1}{c|}{\cellcolor[HTML]{F8696B}50.7} & \multicolumn{1}{c|}{\cellcolor[HTML]{63BE7B}8.3} & \multicolumn{1}{c|}{\cellcolor[HTML]{F8696B}9.3} & \multicolumn{1}{c|}{\cellcolor[HTML]{B9D780}58.5} & \cellcolor[HTML]{63BE7B}8.3 \\
			\multicolumn{1}{l|}{Offline} & \multicolumn{1}{c|}{} & \multicolumn{1}{c|}{39.7} & \multicolumn{1}{c|}{43.8} & \multicolumn{1}{c|}{17.4} & \multicolumn{1}{c|}{52.3} & \multicolumn{1}{c|}{52.4} & \multicolumn{1}{c|}{13.0} & \multicolumn{1}{c|}{59.8} & \multicolumn{1}{c|}{62.9} & \multicolumn{1}{c|}{21.1} & 56.8 & \multicolumn{1}{c|}{60.3} & 56.8 \\ \bottomrule
		\end{tabular}%
	}
	\caption{Results of BiSeNet V2 \cite{yu2021bisenet} on \textit{CS} $\rightarrow$ \textit{ACDC} in mIoU (\%) for each subset of ACDC using different pre-training and augmentations strategies. Compared to DeepLabV3+, BiSeNet V2 is more affected by Forgetting.}
	\label{tab:pre-train_bise}
\end{table*}

\begin{table*}
	
	\resizebox{\textwidth}{!}{%
		\begin{tabular}{@{}lccccccccccccc}
			\multicolumn{14}{c}{\textbf{HRNetV2-W48}} \\ \midrule
			\multicolumn{1}{c|}{} & \multicolumn{1}{c|}{\textbf{Cityscapes}} & \multicolumn{3}{c|}{\textbf{Night}} & \multicolumn{3}{c|}{\textbf{Rain}} & \multicolumn{3}{c|}{\textbf{Fog}} & \multicolumn{3}{c}{\textbf{Snow}} \\
			\multicolumn{1}{c|}{} & \multicolumn{1}{c|}{\textit{Test}} & \textit{Zero} & \textit{Test} & \multicolumn{1}{c|}{\textit{}} & \textit{Zero} & \multicolumn{1}{c|}{\textit{Test}} & \multicolumn{1}{c|}{\textit{}} & \textit{Zero} & \textit{Test} & \multicolumn{1}{c|}{\textit{}} & \textit{Zero} & \textit{Test} & \textit{} \\
			\multicolumn{1}{c|}{\multirow{-3}{*}{\textbf{Method}}} & \multicolumn{1}{c|}{\textit{mIoU}} & \textit{Shot} & \textit{mIoU} & \multicolumn{1}{c|}{\textit{Forgetting}} & \textit{Shot} & \multicolumn{1}{c|}{\textit{mIoU}} & \multicolumn{1}{c|}{\textit{Forgetting}} & \textit{Shot} & \textit{mIoU} & \multicolumn{1}{c|}{\textit{Forgetting}} & \textit{Shot} & \textit{mIoU} & \textit{Forgetting} \\ \midrule
			\multicolumn{1}{l|}{FT} & \multicolumn{1}{c|}{\cellcolor[HTML]{FFEB84}70.7} & \cellcolor[HTML]{F8696B}6.1 & \cellcolor[HTML]{F8716C}42.1 & \multicolumn{1}{c|}{\cellcolor[HTML]{F8696B}38.0} & \cellcolor[HTML]{F8696B}22.8 & \multicolumn{1}{c|}{\cellcolor[HTML]{B4D680}59.7} & \multicolumn{1}{c|}{\cellcolor[HTML]{F8696B}37.2} & \cellcolor[HTML]{F8696B}19.9 & \cellcolor[HTML]{EDE683}67.0 & \multicolumn{1}{c|}{\cellcolor[HTML]{F8696B}36.2} & \cellcolor[HTML]{F8696B}15.7 & \cellcolor[HTML]{A9D27F}62.7 & \cellcolor[HTML]{FA8871}44.1 \\
			\multicolumn{1}{l|}{AutoAlb.} & \multicolumn{1}{c|}{\cellcolor[HTML]{63BE7B}72.4} & \cellcolor[HTML]{63BE7B}19.6 & \cellcolor[HTML]{FFEB84}44.8 & \multicolumn{1}{c|}{\cellcolor[HTML]{FB9774}33.1} & \cellcolor[HTML]{80C77D}43.0 & \multicolumn{1}{c|}{\cellcolor[HTML]{FBAE78}58.1} & \multicolumn{1}{c|}{\cellcolor[HTML]{F7E883}12.6} & \cellcolor[HTML]{6BC17C}55.4 & \cellcolor[HTML]{96CD7E}68.2 & \multicolumn{1}{c|}{\cellcolor[HTML]{F2E783}15.2} & \cellcolor[HTML]{7FC67D}37.5 & \cellcolor[HTML]{FEE683}61.8 & \cellcolor[HTML]{F7E883}20.4 \\
			\multicolumn{1}{l|}{Distort} & \multicolumn{1}{c|}{\cellcolor[HTML]{FDCD7E}70.4} & \cellcolor[HTML]{9CCF7F}15.7 & \cellcolor[HTML]{FFEB84}44.8 & \multicolumn{1}{c|}{\cellcolor[HTML]{E6E382}21.7} & \cellcolor[HTML]{D2DE82}33.3 & \multicolumn{1}{c|}{\cellcolor[HTML]{FEE182}58.9} & \multicolumn{1}{c|}{\cellcolor[HTML]{FEEA83}13.0} & \cellcolor[HTML]{C3DA81}38.6 & \cellcolor[HTML]{F8696B}64.3 & \multicolumn{1}{c|}{\cellcolor[HTML]{C3D980}11.7} & \cellcolor[HTML]{DDE283}24.5 & \cellcolor[HTML]{92CC7E}62.9 & \cellcolor[HTML]{E0E282}18.1 \\
			\multicolumn{1}{l|}{Gaus} & \multicolumn{1}{c|}{\cellcolor[HTML]{F8696B}69.4} & \cellcolor[HTML]{FCBA7A}7.8 & \cellcolor[HTML]{DEE283}45.1 & \multicolumn{1}{c|}{\cellcolor[HTML]{FDBF7C}28.8} & \cellcolor[HTML]{FA8E72}24.3 & \multicolumn{1}{c|}{\cellcolor[HTML]{C0D981}59.6} & \multicolumn{1}{c|}{\cellcolor[HTML]{FA8370}32.4} & \cellcolor[HTML]{FCBC7B}24.5 & \cellcolor[HTML]{F5E884}66.9 & \multicolumn{1}{c|}{\cellcolor[HTML]{FCA877}26.6} & \cellcolor[HTML]{F8696B}15.7 & \cellcolor[HTML]{FEDF81}61.6 & \cellcolor[HTML]{FB9674}40.8 \\
			\multicolumn{1}{l|}{ImageNet} & \multicolumn{1}{c|}{\cellcolor[HTML]{DBE182}71.1} & \cellcolor[HTML]{FA8F72}6.9 & \cellcolor[HTML]{63BE7B}46.2 & \multicolumn{1}{c|}{\cellcolor[HTML]{FED780}26.2} & \cellcolor[HTML]{FCB97A}26.0 & \multicolumn{1}{c|}{\cellcolor[HTML]{FDCE7E}58.6} & \multicolumn{1}{c|}{\cellcolor[HTML]{FA8671}31.9} & \cellcolor[HTML]{FDD780}26.0 & \cellcolor[HTML]{FDCD7E}66.2 & \multicolumn{1}{c|}{\cellcolor[HTML]{FCAD78}25.8} & \cellcolor[HTML]{FFEB84}19.8 & \cellcolor[HTML]{FCB77A}60.4 & \cellcolor[HTML]{F8696B}51.0 \\ \hline
			\multicolumn{1}{l|}{CN} & \multicolumn{1}{c|}{\cellcolor[HTML]{FDD680}70.5} & \cellcolor[HTML]{F1E784}9.8 & \cellcolor[HTML]{F8696B}41.9 & \multicolumn{1}{c|}{\cellcolor[HTML]{B2D47F}17.0} & \cellcolor[HTML]{EFE784}29.9 & \multicolumn{1}{c|}{\cellcolor[HTML]{F8696B}57.0} & \multicolumn{1}{c|}{\cellcolor[HTML]{FFEB84}13.1} & \cellcolor[HTML]{FAEA84}28.1 & \cellcolor[HTML]{FCBD7B}65.9 & \multicolumn{1}{c|}{\cellcolor[HTML]{FFE583}17.1} & \cellcolor[HTML]{FEE983}19.7 & \cellcolor[HTML]{F8696B}58.0 & \cellcolor[HTML]{FFE884}21.8 \\
			\multicolumn{1}{l|}{Combined} & \multicolumn{1}{c|}{\cellcolor[HTML]{9BCE7F}71.8} & \cellcolor[HTML]{7FC67D}17.7 & \cellcolor[HTML]{F8696B}41.9 & \multicolumn{1}{c|}{\cellcolor[HTML]{69BF7B}10.4} & \cellcolor[HTML]{63BE7B}46.3 & \multicolumn{1}{c|}{\cellcolor[HTML]{63BE7B}60.4} & \multicolumn{1}{c|}{\cellcolor[HTML]{C3D980}9.3} & \cellcolor[HTML]{63BE7B}56.9 & \cellcolor[HTML]{FEE382}66.6 & \multicolumn{1}{c|}{\cellcolor[HTML]{BBD780}11.1} & \cellcolor[HTML]{63BE7B}41.3 & \cellcolor[HTML]{EEE784}62.1 & \cellcolor[HTML]{9ACD7E}11.3 \\
			\multicolumn{1}{l|}{Replay} & \multicolumn{1}{c|}{\cellcolor[HTML]{FFEB84}70.7} & \cellcolor[HTML]{F8696B}6.1 & \cellcolor[HTML]{D3DF82}45.2 & \multicolumn{1}{c|}{\cellcolor[HTML]{63BE7B}9.8} & \cellcolor[HTML]{F8696B}22.8 & \multicolumn{1}{c|}{\cellcolor[HTML]{EEE683}59.2} & \multicolumn{1}{c|}{\cellcolor[HTML]{63BE7B}3.3} & \cellcolor[HTML]{F8696B}19.9 & \cellcolor[HTML]{63BE7B}68.9 & \multicolumn{1}{c|}{\cellcolor[HTML]{63BE7B}4.4} & \cellcolor[HTML]{F8696B}15.7 & \cellcolor[HTML]{63BE7B}63.3 & \cellcolor[HTML]{63BE7B}5.9 \\
			\multicolumn{1}{l|}{Offline} & \multicolumn{1}{c|}{} & 44.8 & 45.6 & \multicolumn{1}{c|}{32.5} & 57.9 & 57.9 & \multicolumn{1}{c|}{2.4} & 62 & 68.8 & \multicolumn{1}{c|}{2.4} & 58.2 & 63.1 & 4.3 \\ \bottomrule
		\end{tabular}%
	}\caption{Results of HRNetv2 \cite{WangSCJDZLMTWLX19} on \textit{CS} $\rightarrow$ \textit{ACDC} in mIoU (\%) for each subset of ACDC using different pre-training and augmentations strategies. HRNetV2 performs similar to DeepLabv3+ on Cityscapes, but overall is more impacted by forgetting. The combination of ImageNet pre-training, AutoAlbum. and Continual Normalization (\textit{Combined}) leads to a significant reduction of forgetting.}
	\label{tab:pre-train_hrnet}
\end{table*}

\begin{table*}
	\resizebox{\textwidth}{!}{%
		\begin{tabular}{@{}lccccccccccccc}
			\multicolumn{14}{c}{\textbf{RTFormer - Base}} \\ \midrule
			\multicolumn{1}{c|}{} & \multicolumn{1}{c|}{\textbf{Cityscapes}} & \multicolumn{3}{c|}{\textbf{Night}} & \multicolumn{3}{c|}{\textbf{Rain}} & \multicolumn{3}{c|}{\textbf{Fog}} & \multicolumn{3}{c}{\textbf{Snow}} \\
			\multicolumn{1}{c|}{} & \multicolumn{1}{c|}{\textit{Test}} & \textit{Zero} & \textit{Test} & \multicolumn{1}{c|}{\textit{}} & \textit{Zero} & \multicolumn{1}{c|}{\textit{Test}} & \multicolumn{1}{c|}{\textit{}} & \textit{Zero} & \textit{Test} & \multicolumn{1}{c|}{\textit{}} & \textit{Zero} & \textit{Test} & \textit{} \\
			\multicolumn{1}{c|}{\multirow{-3}{*}{\textbf{Method}}} & \multicolumn{1}{c|}{\textit{mIoU}} & \textit{Shot} & \textit{mIoU} & \multicolumn{1}{c|}{\textit{Forgetting}} & \textit{Shot} & \multicolumn{1}{c|}{\textit{mIoU}} & \multicolumn{1}{c|}{\textit{Forgetting}} & \textit{Shot} & \textit{mIoU} & \multicolumn{1}{c|}{\textit{Forgetting}} & \textit{Shot} & \textit{mIoU} & \textit{Forgetting} \\ \midrule
			\multicolumn{1}{l|}{FT} & \multicolumn{1}{c|}{\cellcolor[HTML]{FEE182}68.8} & \cellcolor[HTML]{F8696B}4.2 & \cellcolor[HTML]{DDE182}42.0 & \multicolumn{1}{c|}{\cellcolor[HTML]{FBA176}24.7} & \cellcolor[HTML]{FFEB84}22.7 & \multicolumn{1}{c|}{\cellcolor[HTML]{FEE582}57.7} & \multicolumn{1}{c|}{\cellcolor[HTML]{F8696B}42.5} & \cellcolor[HTML]{FDD57F}19.4 & \cellcolor[HTML]{E3E383}65.2 & \multicolumn{1}{c|}{\cellcolor[HTML]{F8696B}32.2} & \cellcolor[HTML]{FDCE7E}13.4 & \cellcolor[HTML]{FAEA84}60.7 & \cellcolor[HTML]{FB9874}43.7 \\
			\multicolumn{1}{l|}{AutoAlb.} & \multicolumn{1}{c|}{\cellcolor[HTML]{FDCE7E}68.5} & \cellcolor[HTML]{92CC7E}13.3 & \cellcolor[HTML]{FEE783}41.4 & \multicolumn{1}{c|}{\cellcolor[HTML]{EBE582}18.2} & \cellcolor[HTML]{89C97E}36.8 & \multicolumn{1}{c|}{\cellcolor[HTML]{FBA075}56.0} & \multicolumn{1}{c|}{\cellcolor[HTML]{F4E883}20.6} & \cellcolor[HTML]{97CD7E}42.4 & \cellcolor[HTML]{FBAC77}61.6 & \multicolumn{1}{c|}{\cellcolor[HTML]{F8E983}18.9} & \cellcolor[HTML]{B0D580}26.5 & \cellcolor[HTML]{F8696B}58.0 & \cellcolor[HTML]{FB9F76}43.1 \\
			\multicolumn{1}{l|}{Distort} & \multicolumn{1}{c|}{\cellcolor[HTML]{63BE7B}70.9} & \cellcolor[HTML]{63BE7B}16.3 & \cellcolor[HTML]{63BE7B}43.4 & \multicolumn{1}{c|}{\cellcolor[HTML]{D1DD81}15.7} & \cellcolor[HTML]{9CCF7F}34.5 & \multicolumn{1}{c|}{\cellcolor[HTML]{A2D07F}58.6} & \multicolumn{1}{c|}{\cellcolor[HTML]{F7E883}21.0} & \cellcolor[HTML]{83C87D}46.4 & \cellcolor[HTML]{63BE7B}67.2 & \multicolumn{1}{c|}{\cellcolor[HTML]{E7E482}17.2} & \cellcolor[HTML]{8DCA7E}31.6 & \cellcolor[HTML]{63BE7B}62.0 & \cellcolor[HTML]{E5E382}31.1 \\
			\multicolumn{1}{l|}{Gaus} & \multicolumn{1}{c|}{\cellcolor[HTML]{F8696B}66.9} & \cellcolor[HTML]{F5E884}6.9 & \cellcolor[HTML]{FED980}40.7 & \multicolumn{1}{c|}{\cellcolor[HTML]{FB8F73}25.8} & \cellcolor[HTML]{F8696B}13.6 & \multicolumn{1}{c|}{\cellcolor[HTML]{EDE683}58.0} & \multicolumn{1}{c|}{\cellcolor[HTML]{F9766E}40.5} & \cellcolor[HTML]{F86B6B}14.4 & \cellcolor[HTML]{FCBC7B}62.4 & \multicolumn{1}{c|}{\cellcolor[HTML]{FCB37A}25.1} & \cellcolor[HTML]{F8696B}7.4 & \cellcolor[HTML]{FEE382}60.5 & \cellcolor[HTML]{F8696B}47.9 \\
			\multicolumn{1}{l|}{ImageNet} & \multicolumn{1}{c|}{\cellcolor[HTML]{6CC17C}70.8} & \cellcolor[HTML]{FCBD7B}5.5 & \cellcolor[HTML]{CBDC81}42.2 & \multicolumn{1}{c|}{\cellcolor[HTML]{F8696B}28.1} & \cellcolor[HTML]{FEE683}22.4 & \multicolumn{1}{c|}{\cellcolor[HTML]{63BE7B}59.1} & \multicolumn{1}{c|}{\cellcolor[HTML]{FA7E6F}39.3} & \cellcolor[HTML]{FBEA84}21.4 & \cellcolor[HTML]{C3DA81}65.7 & \multicolumn{1}{c|}{\cellcolor[HTML]{FB9D75}27.2} & \cellcolor[HTML]{F4E884}16.7 & \cellcolor[HTML]{6FC27C}61.9 & \cellcolor[HTML]{FECE7F}38.9 \\ \hline
			\multicolumn{1}{l|}{CN} & \multicolumn{1}{c|}{\cellcolor[HTML]{F3E884}69.1} & \cellcolor[HTML]{F98971}4.7 & \cellcolor[HTML]{F8696B}35.0 & \multicolumn{1}{c|}{\cellcolor[HTML]{FFE283}20.7} & \cellcolor[HTML]{FA9072}16.4 & \multicolumn{1}{c|}{\cellcolor[HTML]{F97D6E}55.1} & \multicolumn{1}{c|}{\cellcolor[HTML]{FFE683}22.8} & \cellcolor[HTML]{F8696B}14.3 & \cellcolor[HTML]{F8696B}58.2 & \multicolumn{1}{c|}{\cellcolor[HTML]{FFE483}20.3} & \cellcolor[HTML]{FBA175}10.7 & \cellcolor[HTML]{FA9072}58.8 & \cellcolor[HTML]{F1E783}33.5 \\
			\multicolumn{1}{l|}{Combined} & \multicolumn{1}{c|}{\cellcolor[HTML]{6CC17C}70.8} & \cellcolor[HTML]{84C87D}14.2 & \cellcolor[HTML]{EEE683}41.8 & \multicolumn{1}{c|}{\cellcolor[HTML]{F8E983}19.5} & \cellcolor[HTML]{63BE7B}41.2 & \multicolumn{1}{c|}{\cellcolor[HTML]{70C27C}59.0} & \multicolumn{1}{c|}{\cellcolor[HTML]{9DCE7E}9.7} & \cellcolor[HTML]{63BE7B}53.1 & \cellcolor[HTML]{D0DE82}65.5 & \multicolumn{1}{c|}{\cellcolor[HTML]{B5D57F}12.0} & \cellcolor[HTML]{63BE7B}37.6 & \cellcolor[HTML]{92CC7E}61.6 & \cellcolor[HTML]{A0CF7E}17.2 \\
			\multicolumn{1}{l|}{Replay} & \multicolumn{1}{c|}{\cellcolor[HTML]{FEE182}68.8} & \cellcolor[HTML]{F8696B}4.2 & \cellcolor[HTML]{FDC97D}39.9 & \multicolumn{1}{c|}{\cellcolor[HTML]{63BE7B}5.1} & \cellcolor[HTML]{FFEB84}22.7 & \multicolumn{1}{c|}{\cellcolor[HTML]{F8696B}54.6} & \multicolumn{1}{c|}{\cellcolor[HTML]{63BE7B}2.3} & \cellcolor[HTML]{FDD57F}19.4 & \cellcolor[HTML]{FEE282}64.3 & \multicolumn{1}{c|}{\cellcolor[HTML]{63BE7B}3.5} & \cellcolor[HTML]{FDCE7E}13.4 & \cellcolor[HTML]{FEE883}60.6 & \cellcolor[HTML]{63BE7B}4.8 \\
			\multicolumn{1}{l|}{Offline} &  & 41.8 & 42.7 & 4.2 & 53.4 & 58.6 & 8.0 & 60.3 & 64.3 & 6.4 & 60.7 & 62.7 & 6.9 \\ \bottomrule
		\end{tabular}%
	}
	\caption{Results of RTFormer \cite{wang2022rtformer} on \textit{CS} $\rightarrow$ \textit{ACDC} in mIoU (\%) for each subset of ACDC using different pre-training and augmentations strategies. }
	
\end{table*}

\section{Comparison of Class- and Domain-Incremental Learning}
\begin{figure}
	\centering
	\includegraphics[width=\columnwidth]{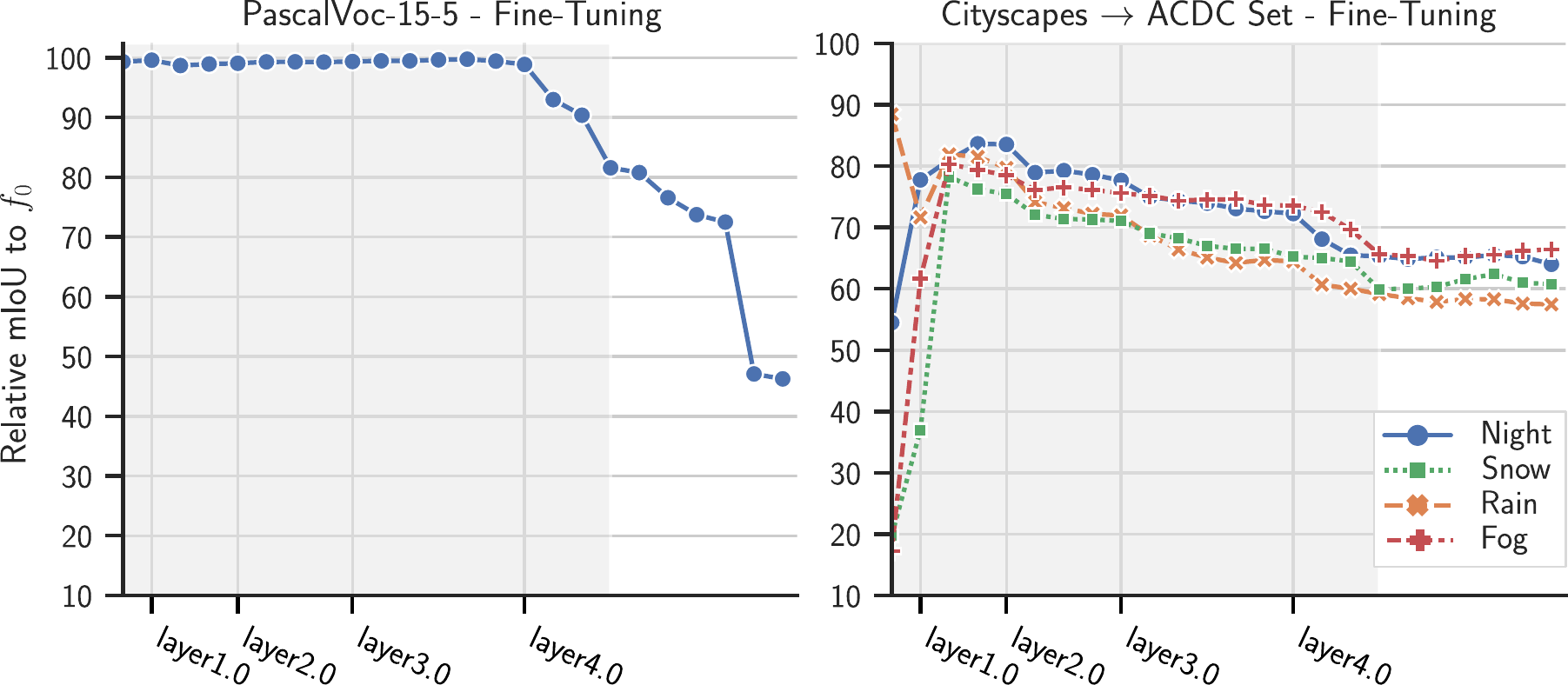}
	\caption{Layer-stitching reveals that during class-incremental learning (PascalVoc-15-5) the encoder layers are mostly stable, only the decoder layers are changing drastically. In domain-incremental learning observe the opposite, early layers show a big discrepancy and later layers do not change as much.}
	\label{fig:layer_stitch_domain_class}
\end{figure}

\cref{fig:layer_stitch_domain_class} shows a comparison of layer stitching for class- and domain-incremental learning. 
In the class-incremental setting, we use PascalVOC2012 \cite{pascal-voc-2012} with the PascalVOC-15-5 split and in the domain-incremental setting, we use the same Cityscapes to \textit{ACDC} setups as before. We see that during class-incremental learning, the encoder layers up until \textit{layer4.0} are not at all affected by activation drift and the representation shift is only affecting late decoder layers. However, in the domain-incremental setting, we see that primarily the first layers are affected by the activation drift and later layers slightly. 

\end{document}